\documentclass{ieeeaccess}
\usepackage{cite}
\usepackage{amsmath,amssymb,amsfonts}
\usepackage{algorithmic}
\usepackage{graphicx}

\usepackage{textcomp}
\usepackage{rotating}
\usepackage{lscape}
\usepackage{caption}
\usepackage{subcaption}
\usepackage{setspace}
\DeclareCaptionLabelSeparator{emdash}{\textemdash}

\def\BibTeX{{\rm B\kern-.05em{\sc i\kern-.025em b}\kern-.08em
    T\kern-.1667em\lower.7ex\hbox{E}\kern-.125emX}}
\begin{document}
\history{Date of publication xxxx 00, 0000, date of current version xxxx 00, 0000.}
\doi{10.1109/ACCESS.2017.DOI}

\title{Explainable AI: current status and future directions}
\author{\uppercase{PRASHANT GOHEL}\authorrefmark{1},
\uppercase{Priyanka Singh \authorrefmark{1}, and Manoranjan Mohanty
}\authorrefmark{2}
}
\address[1]{DA-IICT, Gandhinagar, Gujarat, India %(e-mail: gohel.prashant334@gmail.com)
}
\address[2]{Centre for Forensic Science, University of Technology Sydney, Australia}

\markboth
{Gohel \headeretal: Explainable AI: current status and future directions}
{Gohel \headeretal: Explainable AI: current status and future directions}

\corresp{Corresponding author: Prashant Gohel (e-mail: 202021003@daiict.ac.in).}

\begin{abstract}
Explainable Artificial Intelligence (XAI) is an emerging area of research in the field of Artificial Intelligence (AI). XAI can explain how AI obtained a particular solution (e.g., classification or object detection) and can also answer other "wh" questions. This explainability is not possible in traditional AI. Explainablity is essential for critical applications, such as defence, health care, law and order, and autonomous driving vehicles, etc, where the \textit{know how} is required for trust and transparency. A number of XAI techniques so far have been purposed for such applications. This paper provides an overview of these techniques from a multimedia (i.e., text, image, audio, and video) point of view. Advantages and shortcomings of these techniques have been discussed, and pointers to some future directions have also been provided. 
\end{abstract}


\titlepgskip=-15pt

\maketitle

\section{Introduction}
\label{sec:introduction}
%\PARstart{T}{his} document is a template for \LaTeX. If you are 
%reading a paper or PDF version of this document, please download the 
%electronic file, trans\_jour.tex, from the IEEE Web site at \underline
%{http://ieeeauthorcenter.ieee.org/create-your-ieee-article/}\break\underline{use-authoring-tools-and-ieee-article-templates/ieee-article-}\break\underline{templates/} so you can use it to prepare your manuscript. If 

%Deep learning based algorithms are performing  cognitively hence there is a surge in deep learning  based AI/ML applications. In recent years, AI/ML based applications have been applied to various aspects of human life like science, business, finance and social networking. This surge is due to advance research in field of deep learning where neurons and  hyper parameters are being trained to carry out particular task. Deep learning algorithms are widely used in autonomous industry, healthcare, image processing and speech technology hence reliability without compromising ethics are one of the prime concern for adaption.

%TALK HOW AI IS BEING USED
In recent years, Artificial Intelligence (AI)-based applications have been used in various aspects of human life, such as science, business, finance and social networking, etc. AI-based algorithms have  been successfully applied to all types of data (text, image, audio, video) in various applications, such as healthcare, defence, law and order, governance, autonomous industry, etc. An AI algorithm can now efficiently solve a classification, regression, clustering, transfer learning or optimizations problem \cite{Qizhe}. This current day AI is mainly limited to a sub-branch known as machine learning (ML). Machine learning provides a computer with a set of examples (aka training data set), and let the computer learn from the example set. Once well trained, the computer can then answer questions related to what it was taught previously. Typically, this traditional AI is a \textit{blackbox} that can answer “yes” and “no” type questions without elaborating how that answer is obtained. 

In many applications, an explanation of how an answer was obtained is crucial for ensuring trust and transparency. An example of one such application is a medical application, where the doctors should be damn sure about a conclusion. They, for example, would like to know how AI decided whether someone is suffering from a disease by analyzing a CT scan image. AI-based systems are not 100\% perfect. An insight of how a result was obtained will therefore not only can induce trustfulness but also can avoid life-threatening errors. In some other applications (e.g., law and order), answers to other "wh" questions (such as "why", "when", "where", etc.) could be required. The traditional AI is unable to answer these "wh" questions. 

\begin{figure}
\includegraphics[scale=0.5]{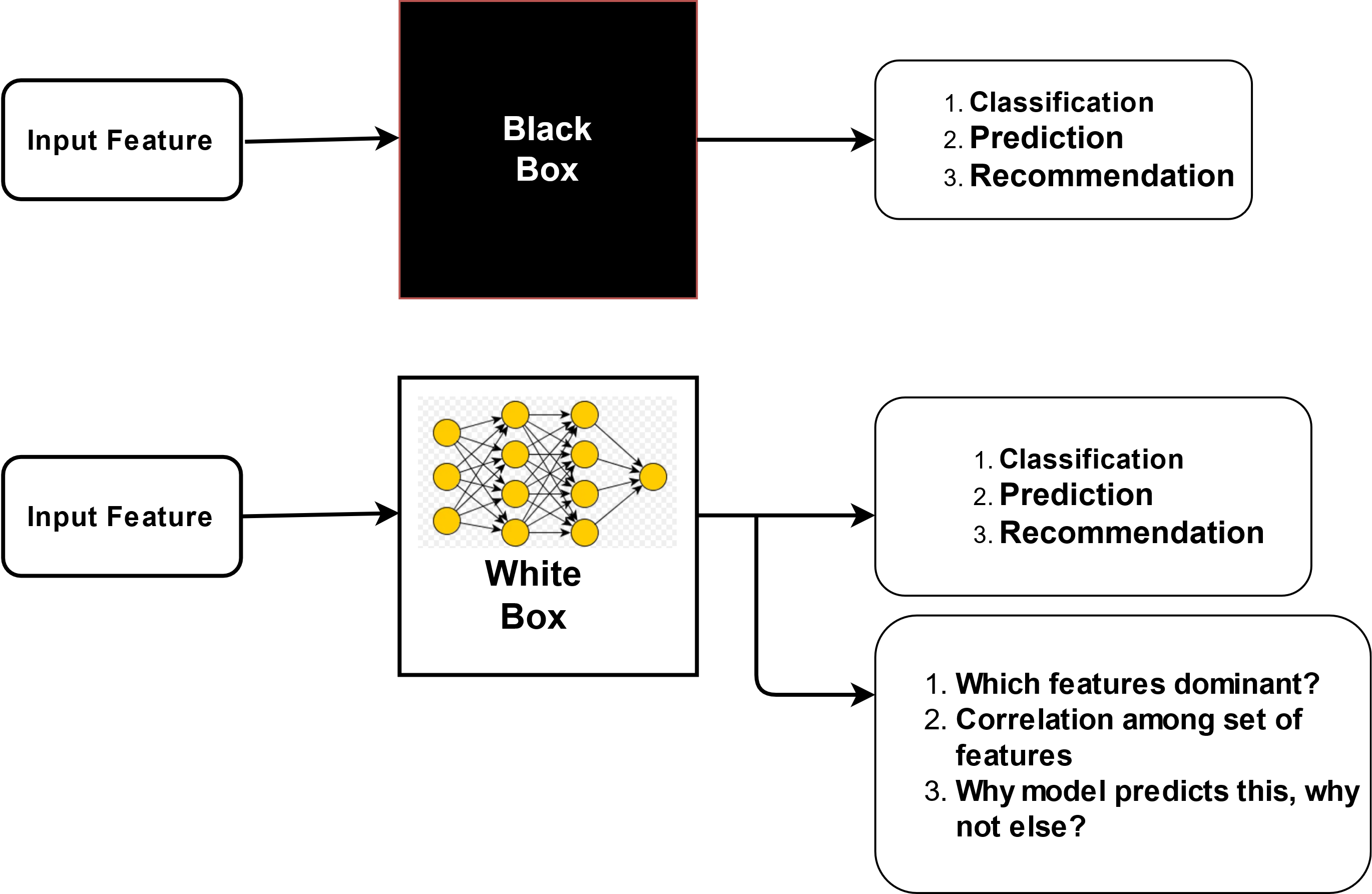}
\caption{AI vs XAI}
\label{fig:transition}
\end{figure}

%Describe Explaianlbe AI
This explainability requirement lead a new area of AI research, know as Explainable AI (XAI). Figure~\ref{fig:transition} shows how XAI can add new dimensions to AI by answering the "wh" questions that were missing in traditional AI. The XAI, therefore, has drawn a great interest from critical applications, such as health care, defence, law and order, etc., where explaining how an answer was obtained (i.e., answers to "wh" questions) is as important as obtaining the answer. In both academia and industry, XAI research, therefore, has become a priority. Although a number of work have been already proposed, more and more work is required to realize the full potential of XAI.

%To provide justifications for results which are produced by deep learning algorithms, requires intrinsic knowledge about how model is trained and how it produces results. Generally end users are not enough qualified to understand this. It is often  observed that ability to provide interpretation is missing to cross human level accuracy. Explanations consists of pixel region for image processing and computer vision related applications and for natural language processing based systems it requires saliency maps of words. Interpretation also consists of values of hyper parameters.

%\subsection{Abbreviations and Acronyms}
%Define abbreviations and acronyms the first time they are used in the text, 
%even after they have already been defined in the abstract. Abbreviations 
%such as IEEE, SI, ac, and dc do not have to be defined. Abbreviations that 
%$incorporate periods should not have spaces: write ``C.N.R.S.,'' not ``C. N. 
%R. S.'' Do not use abbreviations in the title unless they are unavoidable 
%(for example, ``IEEE'' in the title of this article).

%WHAT ARE WE DOING
In this paper, we survey existing research on XAI from multimedia (text, image, audio, and video) point of view. Since each media is different than the other (i.e., image is different than video) in some sense, an XAI method applicable to one media may not be effective for another. We group the proposed XAI methods for each media, point out their advantages and disadvantages, and provide pointers to some future works. We believe that our classification of XAI methods will provide a guide and inspirations to future research in multi-modal applications (for example, XAI for defence where AI-based solutions are required for image, text, audio, etc.).

The rest of the paper is organized as follows. Section \ref{sec:blackbox} discusses why classical blackbox AI. In Section \ref{sec:xai_tool}, we introduce XAI by discussing its scopes, objectives, and various tools proposed to realize explainability. This section also provides a classification tree outlining various XAI methods proposed for multimedia data. These methods are elaborated in the remaining sections. Section \ref{sec:xai tree} explains classification tree for XAI techniques, where transparent and post hoc texhniques are explained.  
%This section also explains how XAI provides more trust oriented results along with debates in ethical community in cyber space. In section \ref{sec:xai_tool}, we have discussed about scopes and objectives of XAI. In this section, we classify different AI We also discussed selected tools of XAI like
Section \ref{sec:image classification} discusses XAI methods applied to image data.
%about LIME, SHAP and counterfactual explanation techniques for explainability with respect to Image data. 
Section \ref{sec:nlp} discusses XAI methods applied to natural language processing (text data). Section \ref{sec:video_sec} and \ref{sec:audio_sec} gives understanding about how XAI works with video and audio type of data. Section \ref{sec:predictive maintenance}  explains about multi modal data like caliberated data from  sensors in CSV format and also explains prevalence of XAI in defence and industrial applications for providing predictive maintenance to reduce cost of production and maintenance.

%In section \ref{sec:nlp} we discusses about explainability and visualization techniques with respect to natural language processing data. 
%Section discusses XAI methods applied to natural language processing (text data).

%Section \ref{sec:healthcare} explains about how XAI helps medical work force to provide better results with justifications. Section \ref{sec:hate speech} explains about how XAI helps to classify hate speech and normal speech with explanations of which word is dominating to classify as a hate or normal. 
%In the following section we discussed about conventional machine learning as known as black box machine learning and key issues with it. Next section is all about usage of XAI as a tool to open black box, objectives of XAI and various approaches of XAI techniques.

%In further stage image classification with various approaches of XAI is explained. XAI and NLP is discussed in next sections with different techniques of explainability and visualization. Explainability methods for health care applications like COVID-19 detection and tumor detection with CT scan image analysis is discussed in next section. Hate speech detection and predictive maintenance with XAI is discussed followed by conclusion and further research scope.

\section{The blackbox AI}
\label{sec:blackbox}

\subsection{Overview}
Blackbox models of machine learning are rapidly being used with a tag of AI-enabled technology for various critical domains of human life. The list of domains varies from socio-economic justice, cyber forensics, criminal justice, etc. But these AI-powered models are lagging to win the trust of naive people because these models are less transparent and less accountable \cite{mashrur}.  For example, there are cases in criminal justice where the AI-enabled justice model release criminals on parole and grants bail. This leads to serious consequences among people and the government \cite{rigano}.

% In section \ref{explainability}, need of explainability is justified with help of case studies. Section \ref{sec:AI and the ‘black box’} explains about usage of XAI. Policy implementation for XAI is explained section \ref{sec:research debates}. Section \ref{sec:key issues} explains about key issues with real time implementation of XAI.

\subsection{Explainability Requirement}
\label{explainability}
“Explainability” is a need and expectation that makes "decision" of intrinsic AI model more transparent. This need will develop rationale approach to implementing action driven by AI and also helpful for end users to understand \cite{robbins}.

In some basic level applications of AI, such as symptoms based  health diagnosis, explainability is straightforward. But in the race of achieving more human level accuracy, researchers and scientists developed more complex algorithms. Neural network and deep learning based applications for decision making is quite elusive and less interpretable \cite{turing} \cite{igami}.

\subsubsection{Case Study 1:  Oil Refinery Assets Reliability: Furnace Flooding Predictions }

\textbf{The Situation}:
Stable combustion is critical for uninterrupted operation of furnace. Due to unidentified factors, if stable combustion is interrupted, it leads to disastrous incident. For consideration of safety, it is highly required that such conditions need to be identified and acted upon. If such preventive measures are not taken, furnace may flood and eventually occurs in explosion. 
Such incident turns into shut down of plant which causes production delay and huge maintenance cost. Unplanned shutdown of industrial plant results in to huge financial loss.

\textbf{The ML/AI Solution}:
A reliable prediction of furnace flooding is required  to alert maintenance staff at least 30 minutes. To develop such kind of predictive maintenance model, it requires to collect calibration from all sensor data including weather and humidity etc. This model should predict flooding prediction at least 20 minutes prior.

\textbf{Why Explainable AI}: 
There could be n number of different factors which are responsible for flooding or shut down of combustion chamber or furnace.
After shutting down the furnace maintenance staff needs to investigate cause of failure. This investigation helps to identify which sensor causes unstability  of continuous operation in furnace.
Making prediction in such a way that helps to identify cause of failure with respect to different parts of combustion chamber. This explainability makes easier to address snag in industrial operation with out wasting much time in investigation.

% \subsubsection{Case Study 1 :  VIDEO TREAT DETECTION }

\subsubsection{Case Study 2:  Video Threat Detection}

\textbf{The Situation}:
In today's age, it is required to secure physical asset and human. A typical solution is to combine security personnel and digital camera for video analytics. Due to human eye limitation. it is not possible to watch every entry point and every video feed at all times. Absolute human supported surveliance may have certain error and threats of miss identification.

\textbf{The ML/AI Solution}:
AI and deep learning based models evaluate video feeds to detect threats. These threats are later flagged for the security personnel. AI based object or face recognition models evaluate video feeds at air port to identify visitors who carry weapon or those are known criminals. These AI model should ignore normal employee or air port staff.

\textbf{Why Explainable AI}:
Due to skewness or  bias of model, it may possible that trained AI model detects innocents visitors or employees due to certain weighted features in training samples. Such kind of incidents raises legality aspects for genuinity of such surveliance systems. Transparency in such systems are one of the crucial factor before framing one person as a criminal or suspect. Company of AI enabled surveliance system are required to provide justification in the court. An individual humiliated and searched publically by security forces leads to several legal consequences on government as well as airport authority.

\subsection{Usage}
\label{sec:AI and the ‘black box’}
AI covers the entire ecosystem of computer-enabled interdisciplinary technologies. 
AI enables a group of technologies to behave more cognitively and context-oriented like human or animal rather than rule-oriented.  AI is all about mimicking the complex cognitive behavior of all living entities on the earth\cite{zhong}.

There are many day-to-day usages of AI-enabled applications
from object recognition, product recommendation in online shopping portals, chatbots for customer service and document processing, etc.  AI is also one of the important tools for medical imaging and diagnosis.
CT scan-based tumor diagnoses are effective and more accurate for certain conditions \cite{zhong}.

AI will have much more applications in the future. 
AI will be a reliable helping hand for doctors to do surgery and diagnoses. Autonomous vehicle driving is one of the upcoming areas where AI will be an important tool. AI can decide while on-road driving for elder people either by taking complete control or by assisting a human driver. For the criminal justice system, AI can make an important decision to declare a person guilty or non guilty \cite{rigano}. AI-enabled decision-making systems provide better support for professionals. Because of the surge in applications of AI in corporate and industry, it becomes a hot topic for ethical concerns. AI usage policies will be decided by government agencies on aspects like privacy, optimization, etc. To make AI more reliable we need to make it more transparent and interpretable. This motivation makes upcoming AI development with aspects of explainability. This area of explainability oriented AI is known as XAI.

%Hence AI-enabled forensic tools are helping hand for investigation agencies.

% As discussed in Fig. \ref{transition}, XAI enables end user to verify factors and its weights behind any decision is derived after training of model. It provides analysis like which factors are dominant and correlations among the features.

%\subsection{The ‘black box’ in policy and research debates}
%\label{sec:research debates}
%With the impact of a huge amount of training data, it is possible to train a model such that it can give more than 99 percent accuracy. But this models very much complex due to its deep learning architecture and hyperparameters. Due to this complexity, it is much difficult to provide transparent and explainable models. Hence this approach is known as  'black box' based AI/ML models.

%These black-box models can be complicated to understand.
%As models are deployed on large-scale systems for policymakers it is a matter of serious concern that whether certain parameters are overweighted to make specific biased decisions.
%Feature importance graphs are crucial for ethical AI policymakers. The essence of feature importance comes with XAI. In the USA, DARPA (Defence Advanced Research Projects Agency)
%emphasize XAI research.

\subsection{Key Issues with Explainable ML}
\label{sec:key issues}
The main reason behind the difficulty to understand and interpret the Black box ML model is either black-box function is quite complicated for a human to understand. Because deep learning-based models are recursive with non-differentiable recursive functions as active functions. Another reason is some functions are proprietary so it is not allowed to expose publicly. There is a belief in XAI researchers that interpretable ML models may reduce the accuracy of prediction and conclusion. Due to this belief, many researchers are now having good expertise in deep learning but not in XAI\cite{rudin}.

Many times explainable AI methods provide justifications that are not aligned with what the original method computes. If explainable methods are computing the same results and predictions as original models then there is no need for an original model. Despite original and XAI models are computing the same predictions there are pretty good chances that both approaches are using a different set of features for making the same predictions. Hence, it is not faithful towards the computation of the black box. It also happens sometimes that the XAI model provides too much extra information which is not relevant to the original inferences of a black-box model.

\begin{figure}[h]
\includegraphics[scale=0.40]{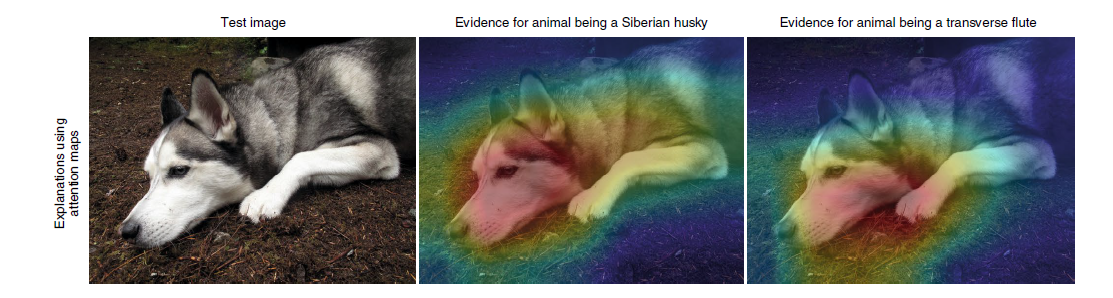}
\caption{Saliency does not explain anything except where the network is looking. We have no idea why this image is labelled as either a dog or a musical
instrument when considering only saliency. The explanations look essentially the same for both classes. Credit: Chaofen Chen, Duke University  \cite{rudin}}.
\label{saliency}
\end{figure}

For image processing domain saliency maps are considered the best tool for image classification. These maps are being useful to determine which part of an image is considered and which is omitted by the model for prediction. But saliency maps are not explaining how different parts of images are contributing to the given prediction. As shown in Fig. \ref{saliency} Saliency maps are not able to demonstrate except where neural network model is focusing.

Consider, for instance, a case where the explanations for multiple (or all) of the classes are identical. This situation would happen often when saliency maps are the explanations, because they tend to highlight edges, and thus provide similar explanations for each class. These explanations could be identical even if the model is always wrong. Then, showing only the explanations for the image’s correct class misleads the user into thinking that the explanation is useful and that the black box is useful, even if neither one of them is.

\begin{figure}[h]
\includegraphics[scale=0.60]{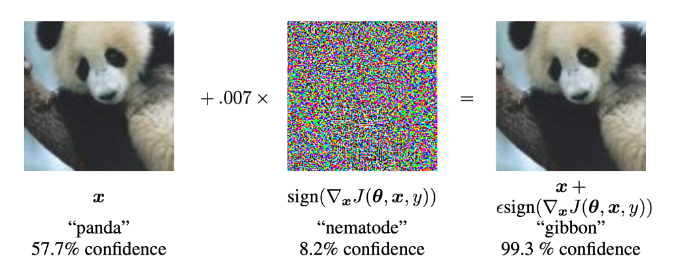}
\caption{ Panda  image  was  tampered  by  adding  some adversarial noise \cite{das}}
\label{pandas}
\end{figure}

Fig. \ref{pandas} illustrates such an example where an image of a  Panda is predicted as a  Gibbon with high confidence after the original Panda image was tampered by adding some adversarial noise.

% Interpretability is  a  desirable  quality  or feature  of  an  algorithm  which  provides  enough  expressive data to understand how the algorithm works. As mentioned in Fig. \ref{Objectives} trustworthiness, transparency, confidence, and informativeness are major objectives of XAI which are must be needed for XAI system validation.

\section{XAI as a tool to open black box}
In section \ref{sec:objectives} objectives like transparency, trust, bias and fairness of XAI are discussed. Section \ref{sec:scopes} provides overview about different scopes of XAI.
\label{sec:xai_tool}

\subsection{Objectives}
\label{sec:objectives}

\begin{figure}[h]
\centering
\includegraphics[scale=0.50]{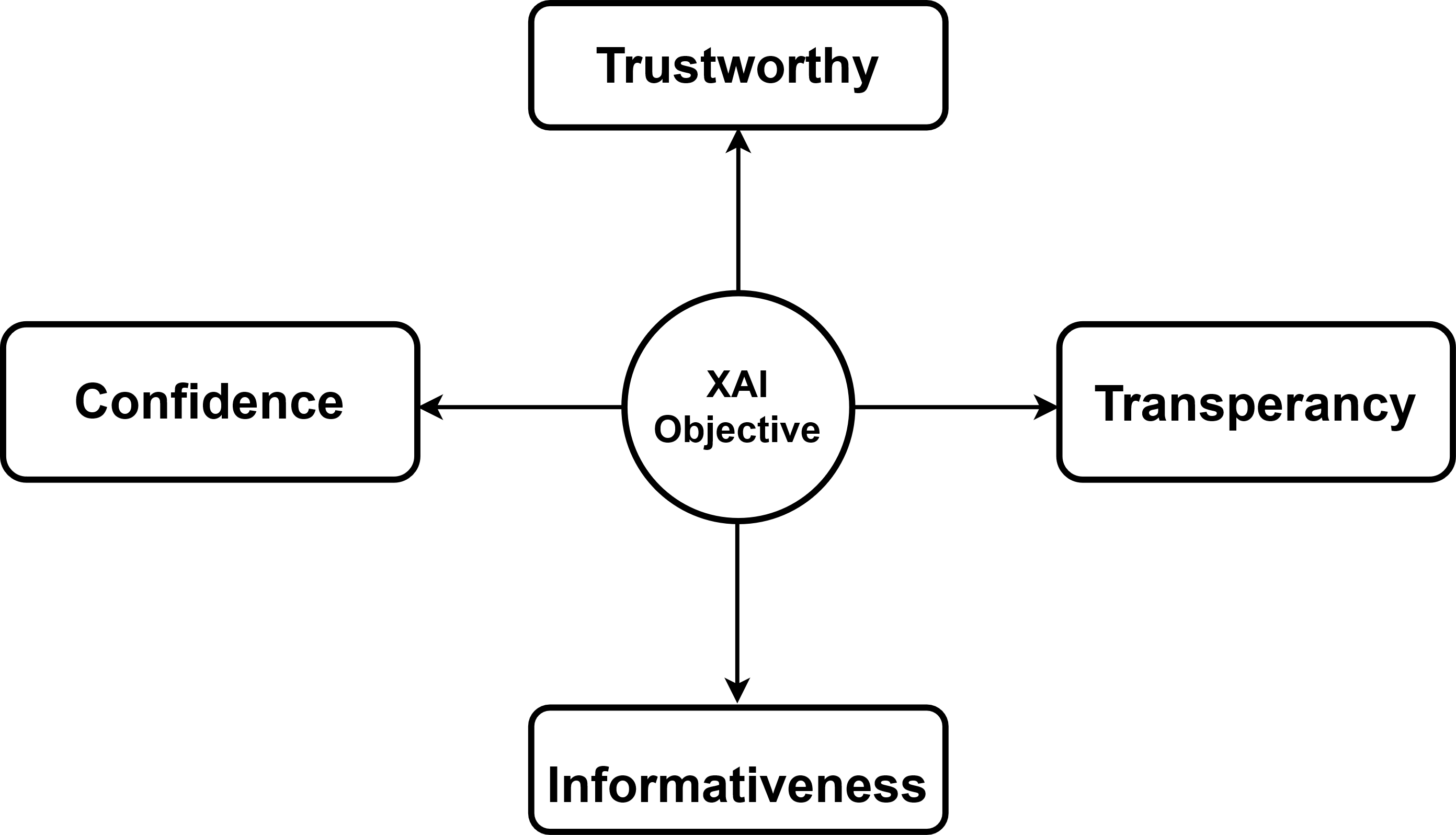}
\caption{Objectives of XAI}
\label{Objectives}
\end{figure}
The main objective of XAI is to answer the "wh" questions related to an obtained answer. For example, XAI should be able to answer "why a particular answer was obtained?", "how a particular answer was obtained", "when a particular AI-based system can fail?" \cite{neer} \cite{garcia} \cite{Mantong} . By doing this, XAI can provide  
trustworthiness, transparency, confidence, and informativeness (Figure \ref{Objectives}).

\subsubsection{Transparency and Informativeness} 
XAI can enhance transparency as well as fairness by providing a justification that can be understood by a layman. The minimum criteria for a transparent AI model are it should be expressive enough to be  human-understandable. 

Transparency is important to assess the performance of the XAI model and its justification. Transparency can assure any false training to model that causes vulnerabilities in the prediction that makes a huge loss in person to the end consumer. False training is possible to tweak the generalization of any AI/ML model that leads to providing unethical benefits to any party unless it is not made transparent.

\subsubsection{Trust and confidence} 
Trust is one of the important factors that makes humans rely on any specific technology. A logical and scientific justification for any prediction and conclusion makes humans favor the prediction or conclusion made by AI/ML algorithms.

\subsubsection{Bias  Understanding  and  Fairness} 
Bias and variance trade-off in AI/ML model makes XAI promote fairness and helps to mitigate bias ( bias-variance trade off)  of prediction at the time of justification or interpretation\cite{dixon}.

\subsection{Scope}
\begin{figure}[h]
\includegraphics[scale=0.5]{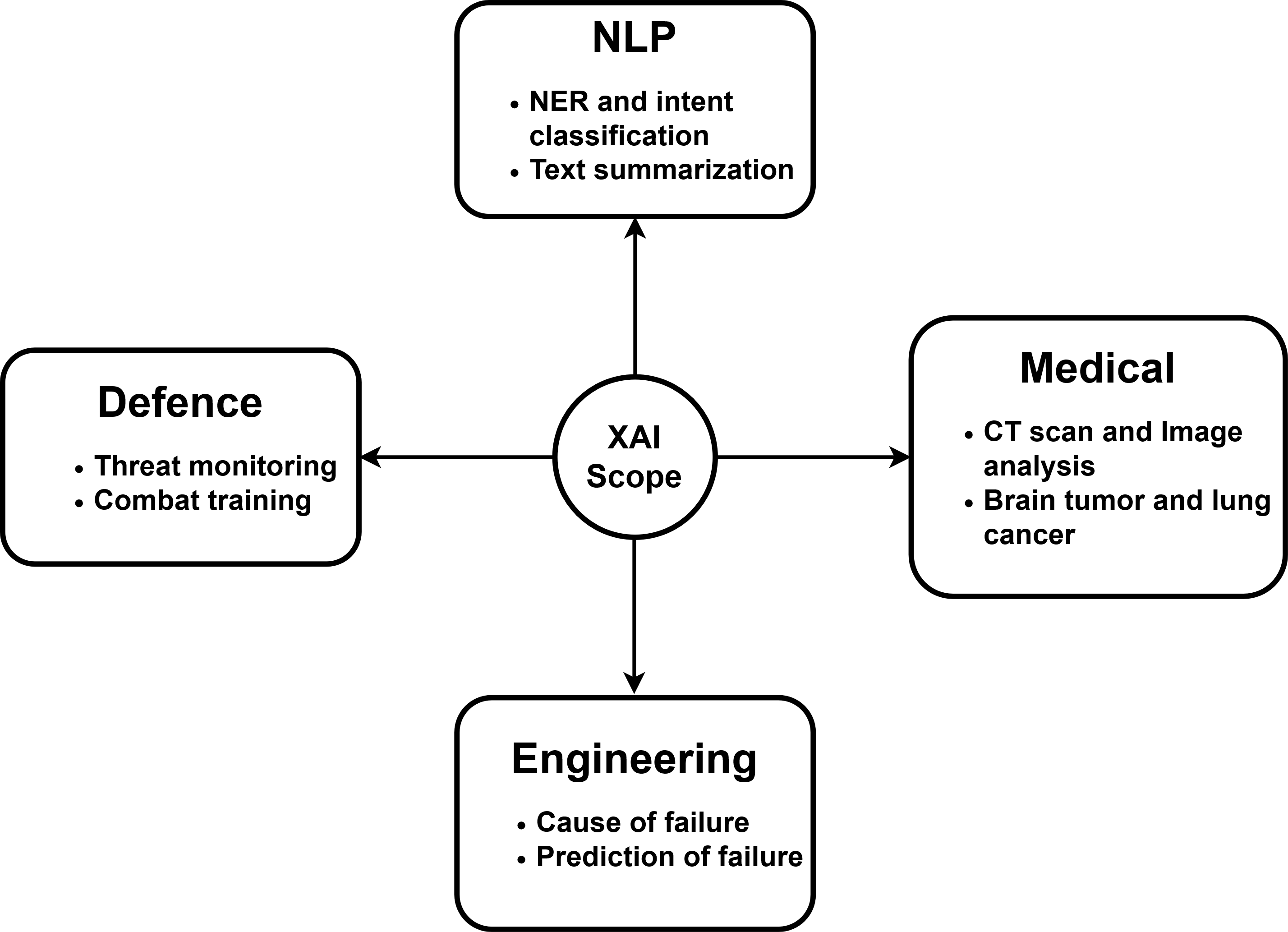}
\caption{Scopes of XAI}
\label{scopes}
\end{figure}
\label{sec:scopes}
% AI has impacted human life in all aspects from medical, finance, lifestyle, transportation, industrial automation, and defense. Neural networks and deep learning's enormous ability to learn things and provide prediction and conclusion makes human life a lot easier to an extent. Scopes of XAI are depicted in Fig. \ref{scopes}. Major scopes of XAI are NLP, medical, engineering, and defense. NLP and engineering comprise banking, finance, digitization, and automation subdomains. 

% Artificial Intelligence (AI) based algorithms, especially using deep neural networks have the intrinsic capability to learn things as a skill without rules and laws behind them. AI is performing real-world tasks with a cognitive approach.  Artificial neurons are having the ability to generalize any sophisticated expertise. Impactful use of AI/ML algorithms in healthcare and vehicles, image processing, and detection in speech surges in our day to day of lives. 

Ideally, scope of XAI can be as broad as scope of AI.  Major scopes are NLP (natural language processing), health care, engineering, and defense. NLP and engineering comprise banking, finance, digitization, and automation. These scopes of XAI are depicted in Figure \ref{scopes}.

1.   Data Protection: European Union and its regulatory body have a 'right to explanation' clause. That makes to enables explanation from XAI algorithms. 

2.     Medical: XAI can diagnose a patient by observing his/her past medical records. Using AI/ML algorithms in the medical image processing domain it is easier for medical experts to diagnose patients with malignant cancer tumors and other lung diseases.

3.     Defense: XAI in defense practices becomes crucial because of automated weapon and surveillance systems. XAI also provides good second-hand support during combat mode training and real-time combat tactics.

4.     Banking: The banking system is one of the biggest financial sectors which affects human life the most. In day-to-day life, there are many fraud transactions and cones by cheaters. Well-trained XAI models can help to investigate fraudulent transactions and help to reduce false positives cases.

\section{Classification Tree}
\label{sec:xai tree}
\begin{figure*}[h]
\includegraphics[scale=0.35]{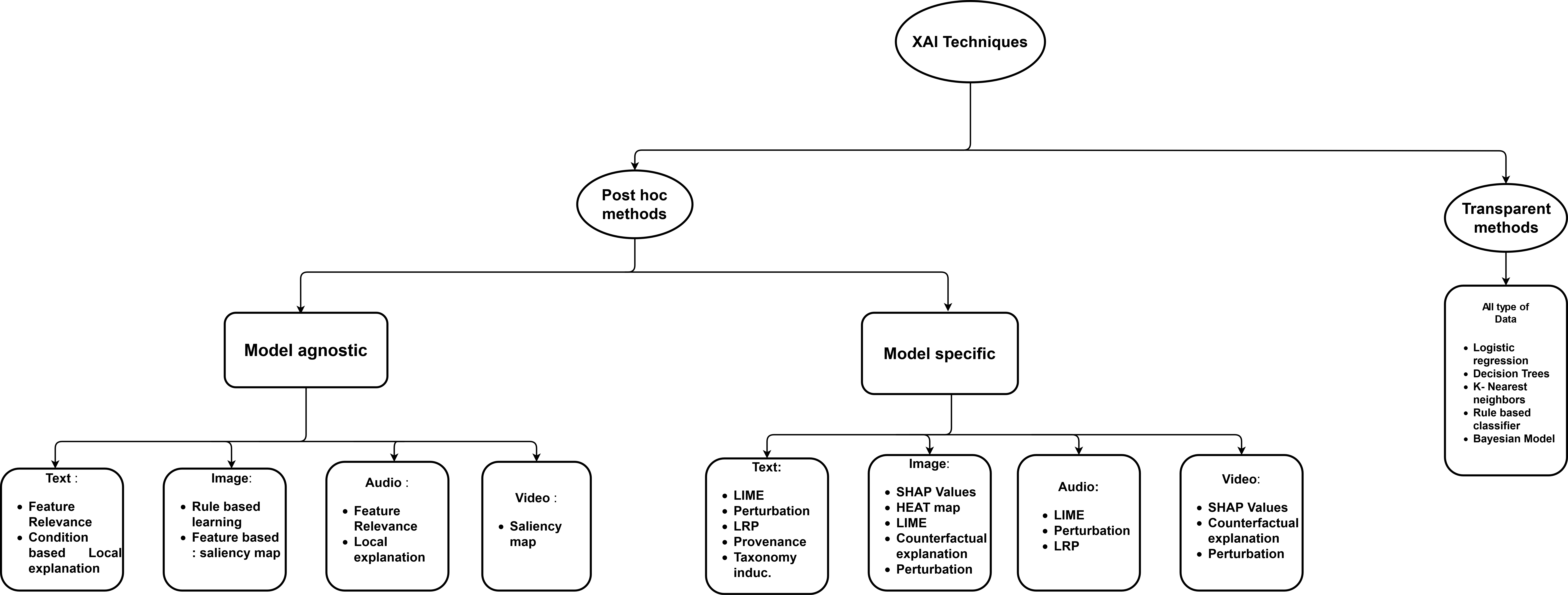}
\caption{XAI classification with respect to type of data}
\label{classification}
\end{figure*}

XAI techniques are classified in two categories of transparent and post-hoc methods. Transparent methods are such methods where the inner working and decision-making process of the model is simple to interpret and represent. Bayesian model, decision trees, linear regression, and fuzzy inference systems are examples of transparent models. Transparent methods are useful where internal feature correlations are not that much complex or linear in nature. Figure \ref{classification} depicts detailed classification of various XAI techniques and approaches with respect to various types of data \cite{Piyawat}.

\subsection{Posthoc methods}
\label{sec:posthoc}

When there is a nonlinear relationship or higher data complexity exists, posthoc methods are useful to interpret model complexity.
In this case, the posthoc approach is a useful tool to explain what the model has learned when it is not following a simple relationship among data and features.

Result-oriented interpretability methods are based on feature summary's statistical and visualization-based presentation. statistical presentation denotes statistics for each feature where the feature's importance is quantified based on its weight in prediction.

A post-hoc XAI method receives a trained and/or tested AI model as input, then generates useful approximations of the model’s inner working and decision logic by producing understandable representations in the form of feature importance scores, rule sets, heat maps, or natural language. Many posthoc methods try to disclose relationships between feature values and outputs of a prediction model, regardless of its internals.  This helps users identify the most important features in an ML task, quantify the importance of features, reproduce decisions made by the black-box model, and identify biases in the model or data.

Some post-hoc methods, such as Local Interpretable Model-agnostic Explanations, extract feature importance scores by perturbing real samples, observing the change in the ML model’s output given the perturbed instances, and building a local simple model that approximates the original model’s behavior in the neighborhood of the original samples. Posthoc methods are further classified in model agnostic and model specific. 
\textbf{Model-specific techniques} supports explainability constraints with respect to learning algorithm and internal structure of given deep learning model. \textbf{model-agnostic} techniques applies pair wise analysis of model inputs and predictions to understand learning mechanism and to generate explanations. 

It is observed that global methods are capable to explain for all data sets while local methods are limited to specific kind of data sets. In contrast,\textbf{model-agnostic} tools can be used for  any AI/ML model. Here pairwise analysis of input and results plays a  key role behind interpretability. In next sections, we have discussed  model specific techniques like feature relevance, condition based explanations,rule based learning and saliency map.

\subsection{Transparent methods}
Transparent method like logistic regression, support vector machine, Bayesian classifier, K nearest neighbour provides justification with local weights of features.
Models falls under this category satisfies three properties named as algorithmic transparency, decomposability and simulatability.

\textbf{Simulatability} stands for simulation of model must be executed by a human. For human enabled simulation complexity of model plays an important role. For an example sparse matrix model is easy to interpret compared dense matrix because sparse matrix model ia easy to justify and visualize by humans.

\textbf{Decomposability} stands for explainability of each aspect of model from input of data to hyper parameters as well as inherent calculations. This characteristics defines behavior of a model and its performance constraints. Complex input features are not readily interpretable. Due to this contraints such models are not belongs to category of transparent model.

\textbf{Algorithmic transparency} defines algorithm level interpretability from input of given data to final decision or classification.
Decision making process should be understood by users with transparency. For an example linear model is deemed transparent because error plot is easy to visualize and interpret. With help of visualization user can understand how model is reacting in different situation.

The transparent model is realized with the following XAI techniques.

\subsubsection{Linear/Logistic Regression}
Logistic Regression (LR) is a transparent  model to predict dependent variable which follows property of binary variable. This method assumes there is a flexible fit between predictors and predicted variables.

For understanding of logistic regression, model it requires audience to have knowledge of regression techniques and its working methodology. Due to this constraints, depending upon type of audience logistic regression falls either in transparent or posthoc methods.Even though logistic regression is the simplest form of supervised classification techniques, its mathematical and statistical concepts are need to be taken care off.

\subsubsection{Decision Trees}
Decision trees is a transparent tool which satisfies transparency in a large context. It is a hierarchical decision making tool. Smaller scale decision trees are easily simulatable.
Increment in number of levels in trees make it more algorithmically transparent but less simulatable.
Due to its poor generalization property, ensembling of trained decision trees are useful to overcome poor generalization property. This modification makes decision tree tool less transparent.

\subsubsection{K-Nearest Neighbors}
KNN (K-Nearest Neighbors) is a voting based tool that predicts class of test sample with help of voting the classes of its k nearest neighbors.
Voting in KNN  depends on distance and similarity between examples. Simple KNN supports tansparency, algorithmic transparency and human centric simulation.
KNN's transparency depends on the features, parameter N and distance function used to measure similarity. Higher value of K impacts simulation of model by human user. Complex distance function restricts decomposability of the model and transparency of algorithmic operation.

\subsubsection{Rule based learning}
Rule based model defines rule to train model. Rule can be defined in the simple conditional if-else form or first order predictive logic. Format of rules depends on type of knowledge base. Rules provides two advantages to this type of model. First, since format of rules are in linguistic terms it is transparent for user to understand. Second, it can handle uncertainty better than classical rule based model \cite{Nicolas}. The number of rules in model improves the performance of model with compromising interpretability and transparency of model. Model with less number of rules can be easily simulated by human.

\subsubsection{Bayesian Model}
Bayesian model are probabilistic model with notion of conditional dependencies between set of dependent and independent variables. Bayesian model is transparent enough for end users who are having knowledge of conditional probability. Bayesian model are enough suitable for all three properties decomposable, algorithmic transparency and human simulation. Complex variable dependency may affect transparency and human simulation for bayesian model.

\label{sec:transparent}

\subsection{Model Specific}
Model specific XAI models are realized using following techniques. 

\subsubsection{Feature Relevance}
It is always important to figure out the most impactful features which are crucial for decision makings. For this feature, importance is introduced. Feature importance shows the impact factor of each feature in derived decisions \cite{Nazneen}. Along with feature importance, correlation among features is also useful for explainability.  In AI-based medical diagnosis model, feature correlation in training data is one of the driving forces for diagnosis.

\subsubsection{Condition based  Explanation}
Condition based explanation is required on the basis of "why", "why despite" and "why given". Some specific observed inputs plays key role to justify prediction. By asking “Why?” oriented  questions, model will provide all possible explanations with set of conditions. This condition set is generated with completeness phenomena. "what if" provides hypothetical reasoning for counterfactual justification. A simple logical model converts user inputs in to the form of constraints based inputs and provide justification that whether constraints are being satisfied in the form of conditions.
 
 \subsubsection{Rule based learning}
 Explainability is required because ML model output is numerical and neural network is too much complex that normal user can not understand the complexity of  hyperparameters and its effect on final prediction.

After getting some insightful understanding of trained model and interpretability of results, suitable approach is to explain derived results to customers and naive users is translation of  those insights into rules such that it can provide full transparency for XAI \cite{Nicolas}.  Once rules are  framed for all possible predictions, It makes even the most complex neural network model transparent.

 \subsubsection{Feature based saliency map}
 Saliency maps are generally used with image processing areas of applications to show what parts of video frames or images are the most significant for derived decision of CNN.
 
 XAI saliency map is a tool which is useful showcase inner working of DNNs. Gradinet computation using back propagation algorithm are used as quantified measures to project  intensity of colours on plane.

\subsection{Model agnostic}
%\label{sec:agnostic vs specific}
%Model-specific approaches are specific single or group of models. These tools rely highly on the working methodology and capacity of a specific or group of models. While model agnostic methods rely on input and output of model rather than internal working of the model.

Model agnostic techniques are also applied for text, image, audio and video. various techniques like LIME, perturbation, LRP, SHAP, provenance and taxonomy inductions, counter factual explanations are applicable on different type of data like text, image, audio and video.

\subsubsection{LIME- Local  Interpretable Model-agnostic Explanations} Model agnosticism specifies the property that LIME is able to provide justification for any type of supervised learning model's prediction. This technique is applicable for any sort of data like image, text and video. This means that LIME is able to handle any supervised learning model and provide justification. 

LIME provides local optimum explanations which computes important features around the vicinity of given particular instance to be explained. 
By default it generates 5000 samples of the feature vector which are following normal distributions. After producing normally distributed samples it finds the target variables for samples whose decisions are explained by LIME.

After obtaining local generated dataset and their predictions it assigns weights to each of the rows how close they are from original samples. Then it uses a feature selection technique like lasso or PCA (Principle Component Analysis) to get significant features.  Detailed discussion about LIME is referred in section \ref{sec:lime}.

LIME has found much success and support in the field of XAI and is implemented for text, image, and tabular data. One noiticeable observation about LIME is that it is applicable and extendable to all significant machine learning domains. In the domain of text processing, embeddings and vectorization of given word or sentence can be considered as a basic unit for sampling. For Image, segmented parts of Image are considered as samples for input.

\subsubsection{Perturbation}
Perturbation helps to generate desired explanation drivers and analyze impact of  perturbed features on the given target. It provides summary of all features for given pertubed results.

In perturbation mechanism local changes are observed on target results and perturbation scores are assigned to all features using LIME or SHAP methods.

Perturbation method is easy to implement and it is not applicable to specific architecture of model. This method can be applied to type of AI/ML model. Disadvantage of perturbation method is, it is computationally expensive if number of features are relatively greater than normal average. As there are more number of features it takes more time to evaluate  combination of all features.

This scenario occurs specifically when dimensions of input are more because number of combinations of all features  grows rapidly. Moreover, this mechanism can underestimate the selected feature's contribution because respective feature reaches saturation level in perturbation such that perturbing them do not have any impact on derived results. 

\subsubsection{LRP: Layer-wise Relevance Propagation}
LRP is useful to unbox complex neural networks. It propogates predictions backward in the neural network. For backward propogations specific rules are designed.

\subsubsection{Provenance and taxonomy induction}
Provenance and taxonomy induction are logical inference based techniques to justify result based on partially derived results. In section  \ref{sec:nlp explainability} it is discussed with detail \cite{Siham}. 
Comprehensive analysis of important XAI techniques is presented in  Table \ref{table:1}.

\begin{table*}[]
\begin{tabular}{|l|l|l|l|}
\hline
Approach &
  Advantages &
  Drawbacks &
  Future Directions \\ \hline
LIME [1][4] &
  \begin{tabular}[c]{@{}l@{}}Plug and play \\ based\end{tabular} &
  \begin{tabular}[c]{@{}l@{}}The resulting explanations are found \\ to   be unstable. The ranking does not \\ account for feature dependence.\end{tabular} &
  \begin{tabular}[c]{@{}l@{}}To reduce local fidelity of justification\\ , Non redundant instance based justifi-\\ cation can be tried.\end{tabular} \\ \hline
SHAP [5][8] &
  \begin{tabular}[c]{@{}l@{}}Optimized for\\ speed up\end{tabular} &
  \begin{tabular}[c]{@{}l@{}}Small perturbations with no change \\   in prediction leads to different \\ explanation.\end{tabular} &
  \begin{tabular}[c]{@{}l@{}}SHAP can be used to define contribution\\ of each feature.\end{tabular} \\ \hline
LRP [0] &
  \begin{tabular}[c]{@{}l@{}}Suitable for\\ neural network\end{tabular} &
  \begin{tabular}[c]{@{}l@{}}Low abstract level explanation\\  with relevance map\end{tabular} &
  \begin{tabular}[c]{@{}l@{}}With layer wise attribution of neural network \\ class discriminavity can be increased\end{tabular} \\ \hline
Heatmap &
  \begin{tabular}[c]{@{}l@{}}Feature importance\\ based presentation\end{tabular} &
  \begin{tabular}[c]{@{}l@{}}Individual pixels are typically \\ not meaningful for \\ humans less interpretable\end{tabular} &
  \begin{tabular}[c]{@{}l@{}}It is a self explainable approach for  image \\ classification but future work required for\\ text-based presentation\end{tabular} \\ \hline
SEDC and   SEDC-T [6][7]&
  \begin{tabular}[c]{@{}l@{}}More human \\ centric explanation\end{tabular} &
  More than one irreducible explanation &
  counter factual analysis for text based classification. \\ \hline
Feature Importance [77][31][32] &
  \begin{tabular}[c]{@{}l@{}}Feature weight based\\ explanation\end{tabular} &
  \begin{tabular}[c]{@{}l@{}}Explanation is limited with respect\\  to only local   features.\\  May drive attention of user\\  towards from important global \\  dependency.\end{tabular} &
  Reduction of local fidelity on prediction \\ \hline
Induction [77] &
  \begin{tabular}[c]{@{}l@{}}More convenient \\ for programming\end{tabular} &
  \begin{tabular}[c]{@{}l@{}}Such techniques assume   that end \\ users can understand \\ specific representations, \\ such as first-order  logic rules and \\ reasoning trees.\end{tabular} &
  Generalization is required for diversified data. \\ \hline
Provenance [77][33] &
  \begin{tabular}[c]{@{}l@{}}Natural language \\ based explanation\end{tabular} &
  \begin{tabular}[c]{@{}l@{}}It is more accessible for lay users \\ but not much   compatible\\  for validation because it is\\  based on natural language.\end{tabular} &
  \begin{tabular}[c]{@{}l@{}}validation of justification is required for \\ more reliability.\end{tabular} \\ \hline
\end{tabular}
\caption{Analysis of various XAI appraoches}
\label{table:1}
\end{table*}

\section{XAI and IMAGE}
\label{sec:image classification}
Explanations in XAI are often categorized into two main aspects. The first category is whether the given explanation is limited to the given conclusion of a model or it describes the entire prediction process which includes training aspects also. The second category differentiates between whether explanation comes directly from the prediction process or it requires posthoc analysis \cite{Adadi} \cite{Karen}.

Popular instance-level explanation methods for image classification such as LIME \cite{marco}, SHAP \cite{SHAP} and LRP \cite{LRP},
typically create feature importance rankings. Although insightful, these methods have clear drawbacks: they do not
determine the optimal explanation size, they do not account for feature dependence, and they are related to only one
prediction class.

\subsection{LIME}
\label{sec:lime}
Local interpretable model-agnostic explanations (LIME), as the name suggests it interprets the model locally and explains the classification of the model in a faithful manner. In LIME, the prediction of the model is used as labels for supervised training to train the XAI model. 
  
\begin{figure}[h]
\centering
\includegraphics[scale=0.40]{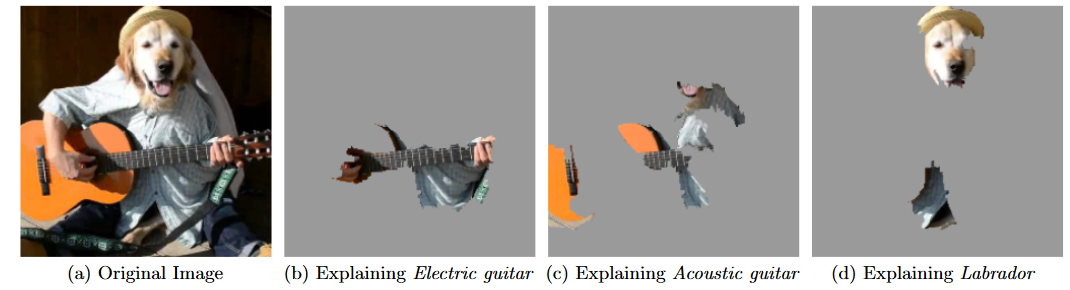}
\caption{ Explaining an image classification prediction made by Google’s Inception neural network.  The top3 classes predicted are “Electric Guitar” (p= 0.32), “Acoustic guitar” (p= 0.24) and “Labrador” (p= 0.21)\cite{lime}}
\label{lime}
\end{figure}  

Sparse linear models are a useful tool to explain LIME-based justification \cite{sparse}. Using a sparse linear model it is possible to highlight important pixels with their weights for a particular respective class as shown in Fig. \ref{lime}. This set of important pixel areas give intuition as to why the model would think that class may be present. As described in figure 5 important pixel-based explanation is given. It interprets the original image as electric guitar, acoustic guitar, and Labrador with respect to the confidence score of 0.32, 0.24, and 0.21.

\subsection{SHAP (SHapley Additive exPlanations)}
\label{sec:shap}
The main objective of SHAP is to understand the prediction of an input A by computing the decision-making contribution of each feature for the classification. SHAP computes Shapley values using coalitional game theory. It is a technique described by Shapley (1953) \cite{lloyd} as an approach for assigning a reward to game players according to their contribution to the game. SHAP assigns each feature an importance value for a particular prediction \cite{shap2}.

The key difference between LIME and SHAP is the process to assign weights to the regression linear model. LIME uses cosine measure between the original and the perturbed image. SHAP, the weights are determined using the Shapley formula. LIME and SHAP  methods have their  drawbacks: they do not determine the optimal explanation size, they do not account for feature dependence, and they are related to only one prediction class.

\subsection{Counterfactual Visual Explanations}
\label{sec:counterfactual}
For human psychology, it is convenient to explain by giving contrastive explanations rather than giving direct explanations to the conclusion or prediction of the machine learning model. We can explain by providing reasons, why only a certain class is selected and why others are rejected.

For explainability, we generally try to provide the explanation on the basis of the selection and rejection of the specific alternatives or outcomes. For given scenario, why only outcome A selected not B. A useful tool to provide such a discriminative explanation is using counterfactuals. We can use counterfactuals to provide reasonably valid arguments at the end of the conclusion by machine learning model which is supported by either deep learning or classical statistical modeling. With the nature of counterfactuals, a certain set of features are defined that can change the decision of the model. If those features are not available then the final conclusion of the model will be changed. It is argued
that they are more likely to comply with recent regulatory developments such as GDPR. The counterfactual approach helps to understand and satisfy three important needs of interpretability: how an interpretation of a model
was made, it provides the scope to tweak with adverse decisions, and gives clues to receive intended results in prediction.

%\begin{figure*}[h]
%\includegraphics[scale=0.5]{XAI_visu.png}
%\caption{ Understanding of explainability}
%\end{figure*}

There is a scenario of classification in classes A, B, and C.
let's say there is a feature set (a1, a2,......, a10) which are relatively required to get prediction A. For Given input result of prediction is class B because feature set (a11,a12,.....a20) is present and  feature set (a1, a2,......,a10) is absent. The approach with the feature set based interpretation leads to smooth convincing in human-critical domains like Crime, Forensic. \cite{dhurandhar}.

\begin{figure}[h]
\includegraphics[scale=0.4]{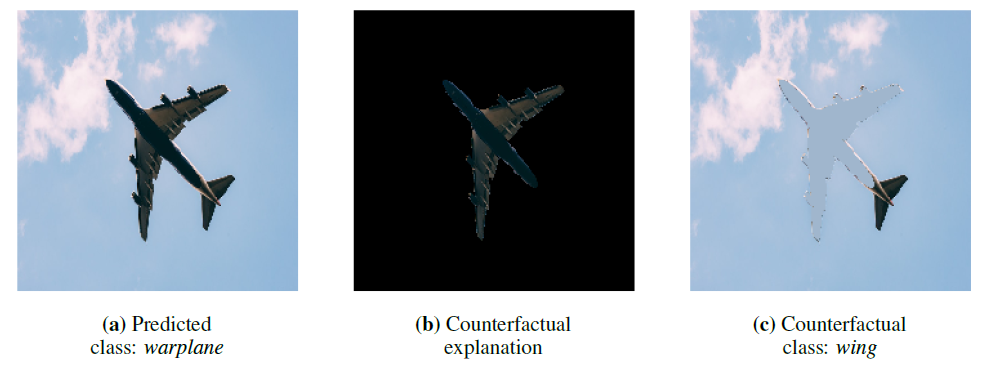}
\caption{Figure a depicts an
image which predicted as warplane by model. Figure b and c shows that by removing body and main wings from plane model predicts tail portion as wing rather than Plane. So figure b is an critically minimum portion in an image to get classified as Warplane.\cite{tom}}
\label{plane}
\end{figure}

This evidence-based approach of counterfactual is known as Search for EviDence Counterfactual for Image Classification(SEDC). Fig. \ref{plane} explains how body of plane i s minimally critically portion to classify as an Image.

As per the research of dhurandhar et al.\cite{dhurandhar} there is a notion of pertinent positive(PP) and pertinent negative(PN). A pertinent positive (PP) is a factor that is minimally required for the justification of the final decision of the model. the pertinent negative is a factor whose absence is minimally required for justifying the conclusion. Figure 6 denotes that the plane body is critical minimum evidence for getting classified as Warplane.

% \usepackage[normalem]{ulem}

The advanced approach is also under research which is known as SEDC-T. Where T stands for predefined target class, not just another class. In SEDC segments images are removed until the predicted class is not changed but in SEDC-T segments are removed from Image until pre-defined class is reached.SEDC-T gives a more detailed explanation of why the image is not predicted as a correct class rather than just explaining the reason behind the prediction of the incorrect class.

\subsection{XAI and Healthcare}
\label{sec:healthcare}
 XAI and healthcare is an effective combo of digital technology. In the trend of AI-based diagnosis systems, trust over AI-based conclusions is a matter of serious concern. Trust is an important factor for the perseverance of AI in the medical and healthcare segment of the digital industry \cite{singh}. If-else diagnosis models are inherently explainable because it consists of feature value sets and will assign a score based on a feature value of an instance case of health diagnosis. If-else-based explainable medical diagnosis systems are well suited for external symptomatic disease diagnosis\cite{yan}. Whether a given deceased person is having asthma or not can be detected by checking whether the symptoms list of the person having what amount of matching criteria with If-else based feature values. For example, if the patient is already having a past history of respiratory illness and cough then there are higher chances of having asthma.
 
 This step-by-step analysis provides a very effective explanation for external symptoms.  To cover a broad spectrum of XAI it is required to make the justification that is independent of the AI model. Such methods are known as model agnostic  XAI methods. LIME (Local  Interpretable  Model-Agnostic Explanation)\cite{marco} is an example of a model agnostic method. LIME is a framework to quantify weights of all factors which are there to make conclusion or prediction. There are other model agnostic XAI techniques also like SHAPLEY\cite{shap2}.  Deep learning is a very important tool for accurate medical diagnosis but its black-box approach for prediction and conclusion makes it restricted for the certain critical area of human medical science.
 \subsubsection{Explainability methods for XAI-Healthcare}
 \label{sec:explainability healthcare}
There are two types of methods for an explanation of medical imaging. One method is based on attribution based and another method is based on perturbation. 
 
\begin{figure*}[h]
\includegraphics[scale=0.6]{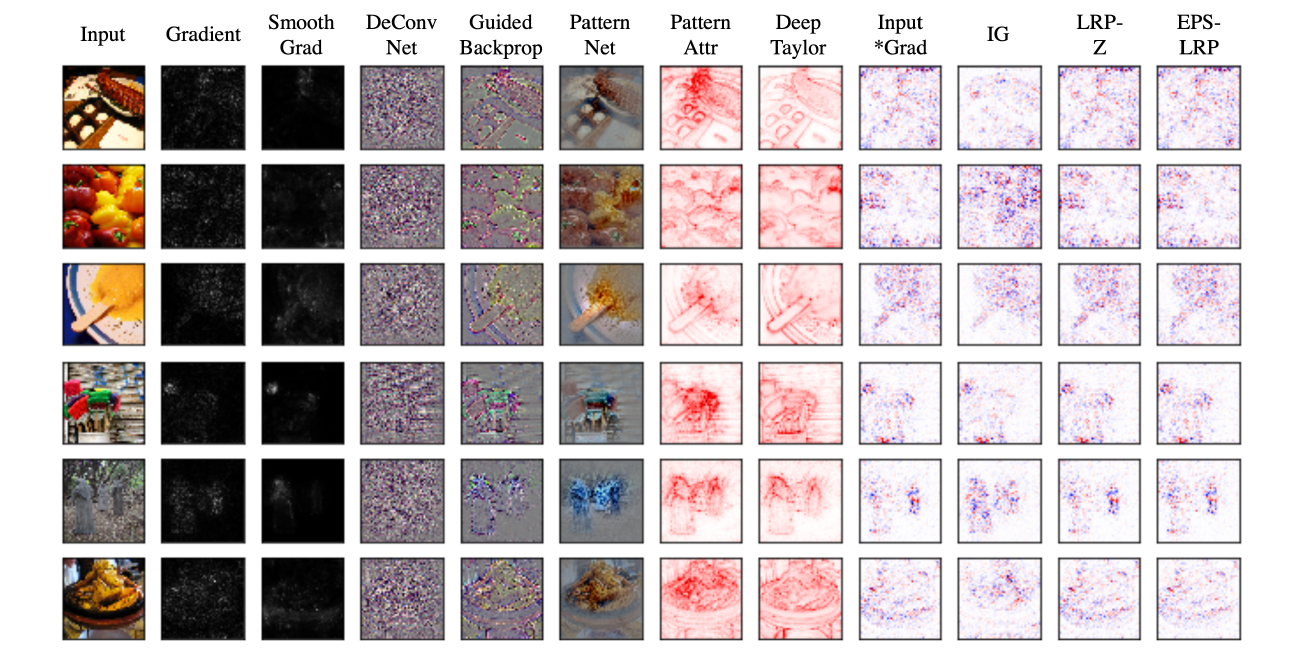}
\caption{Attributions of VGG-16 with images from Imagenet using the methods implemented in \cite{singh}}
\label{attrib}
\end{figure*}

 \textbf{Attribution} LIME is an attribution-based approach for medical image diagnosis. In the attribution-based methods, one needs to determine the contribution and weight of each feature. The success of attribution-based explanation is based on the generality of assigned weights for a given prediction or conclusion at the end of the model. Heat maps are an example of attribution maps. Fig. \ref{attrib} explains how various feature set with respect to different kernels in VGG 16 demonstrates heat map feature weights.
 
 DeepTaylor \cite{taylor} has provided an approach to generating specific positive evidence for a given prediction. The deepTaylor approach of XAI is useful for justifying CNN-based classification. It explains without changing underlying architecture, this property makes it an effective XAI tool. DeepExplain provides a unified framework using  gradient and perturbation-based attribution methods \cite{ancona} \cite{goire}. 
 
DeepLIFT (Deep Learning Important FeaTures) is a technique based on decomposing the prediction of a neural network for specific input. The entire backpropagation process is observed along with observation of weight and bias on each neuron on every layer of the entire architecture. Based on a variety of weights on neurons specific scores are assigned to each feature of input \cite{avanti}.

\textbf{Perturbation} In this approach, input features are getting changed to observe the impact on final prediction at the end of the last layer in the neural network. Perturbation can be achieved by masking or editing certain input features and observations are recorded as model start training using forward pass and backward pass. This is similar to the sensitivity analysis performed in parametric control system models \cite{gilpin} \cite{fernandez}.

The sensitivity of each feature based on input variation is recorded. This continuous observation makes the XAI practitioner justify different predictions at the end of the neural network. Rank assignment to various features is similar to deep explain. one drawback of deeplift approach is computationally expensive. After each forward and backward pass of a number of iterations, observation of sensitivity with respect to features is recorded. Occlusion is an important technique for extracting important features from given Image\cite{erico} \cite{liam}. It is a straightforward model to perform a model agnostic approach that explores the latent feature ranking of a model. All pixels occlusion is computationally expensive, hence 3 x 3 and 10 x 10 tiles are generally useful for occlusion \cite{doshi} \cite{sebastian}.  Trade-off is there with respect to the size of tiles and accuracy.

\subsubsection{XAI for Health care applications}
 \label{sec:health care applications}
 \textbf{Brain Imaging} CNN is a tool for accurate Image classification. Features-based classification of Alzheimer's using CNN gives robust \cite{eitel} classification and accuracy. 

Using post hoc analysis we can understand there is a certain amount of overfitting is available due to certain features. In post model analysis it is possible to tweak certain hyperparameters so that more accurate results are possible to extract. Methods like  Guided backpropagation (GBP), LRP, and DeepShap are useful for brain imaging and classification \cite{pereira}.  

During surge of COVID 19 pandemic AI along with explainability plays an important role in covid 19 diagnosis. Major steps are depicted in Fig. \ref{covid}.

\begin{itemize}
  \item Extraction of lung information from chest CT scan.
  \item Classification of  CT scans in the category of covid positive and negative using convolution.
  \item Localization of lung symptoms like ground glass and crazy paving in CT scans.
  \item  Provide well-documented justification \cite{matteo}.
\end{itemize}

\begin{figure}[h]
\includegraphics[scale=0.55]{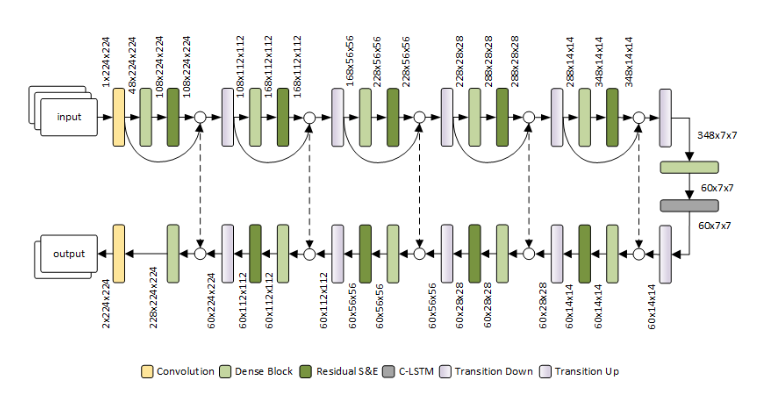}
\caption{The proposed segmentation architecture, consisting of a down sampling path (top)and an up sampling path (bottom), interconnected by skip connections and by the bottleneck layer.\cite{matteo}}
\label{segmentation}
\end{figure}

Fig. \ref{segmentation} explains neural network architecture of covid 19 detection model to test presence of covid 19 by analysing CT scan of lung.  The sequence of 3 consecutive slices  (224×224) lungs CT scan,  which are fed in pipeline individually and combined through a  convolutional  LSTM  layer. The architecture of convolutional LSTM is described in figure 9. The resulting feature maps are then processed with downsampling. Downsampling generates five sequences of dense blocks and then squeezing-excitation is performed. At last, max-pooling operation is performed. At the end, six channel segmentation is generated for lobs and nonlungs area of CT scan as shown in Fig. \ref{covid}.

\begin{figure}[h]
\includegraphics[scale=0.50]{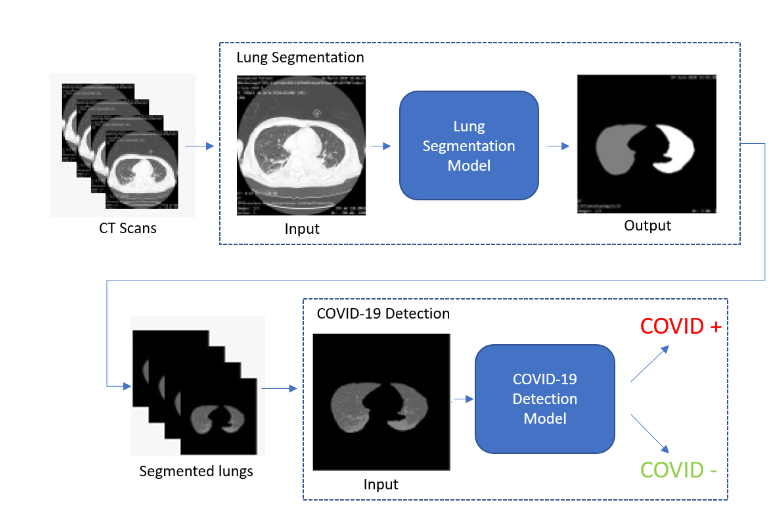}
\caption{Overview of the COVID-19 detection approach for CT scan classification as either COVID-19 positive or negative.\cite{matteo}}
\label{covid}
\end{figure}
.

\section{ XAI and TEXT}
\label{sec:nlp}
In general, Natural language processing (NLP) systems are having inherent explainability. NLP major applications using machine learning are sentiment analysis, hate speech detection, text summary generation \cite{Nikos} \cite{Attention}. For all these applications machine learning models are used like decision trees, sequential modeling, logistic regression, a bag of words, skip grams \cite{Martin} \cite{founta} \cite{glorot}.  Due to the recent advancement in word embeddings, It gives more efficiency to black-box-based inferences and conclusions \cite{james}. One drawback of this increased efficiency is these models are less interpretative and less explainable. Digital ethics are a big concern as far as reliability is concerned over black-box systems. Hence Explainable AI makes more sense for NLP-based applications of AI and Deep Learning \cite{Robert} \cite{Sofia}. Figure 7 consists of various explanation techniques for NLP like a saliency heat map, saliency highlights, declarative explanation, and natural language inference.

\subsection{Explainability Techniques for NLP}
\label{sec:nlp explainability}
There are five different techniques that are useful for providing mathematical justification for the conclusion and classification of AI-based NLP system \cite{Reid} \cite{pelekis}.

\textbf{Feature importance:} This technique is based on the weight of various features based on feature engineering concepts. Specific scores are going to be assigned to individual features based on their contribution to the final prediction. This technique is based on various features of NLP. Some features are handcrafted or annotated, which are extracted manually by feature engineering \cite{Vijay}, lexical features based on tokens and n-gram \cite{Ashish}, or latent features using LDA \cite{Prihatini} and  Attention mechanism \cite{Bahdanau}. Text-based features are more convenient for the human to understand just because they in form of a readable text. There are certain disadvantages also with hand-crafted features due to their local optimum derivations.

\textbf{Example driven}: In an example-driven approach, examples of text being provided are in favor of the final conclusion and some are against the final conclusion. This approach leads to an instance and label base justification for given prediction and conclusion. They are similar in the essence of the neighborhood or clustering-based approach.
 
\textbf{Provenance}: This approach is based on the reasoning steps. It is thoroughly validated approach for reasoning based derived justification. In this approach, final result is derivation from series
of reasoning steps. This is best technique for automatic question answer explanation \cite{abuj}.

\subsection{Visualization Techniques for NLP}
\label{sec:nlp visualization}
Presentation is a crucial segment of XAI as far as justification is concerned for a naive person. There are many ways for a visualization based on chosen XAI approach or technique. For an effective attention-based mechanism that gives weightage to different features, the saliency map-based technique is an important tool that demonstrates scores of individual features \cite{mahnaz} \cite{Jiwei}. Here, we have provided a detailed description of different visualization techniques in fig.6.

\textbf{Saliency:} There is a strong correlation between feature score-based justification and saliency-based visual presentation. There is many research-based demonstrations where saliency-based visualization technique is chosen. Saliency-based visualizations are popular because they present visually perceptive explanations and can be easily understood by different types of end-users \cite{Nina}.

\textbf{Raw declarative representations:} This technique is based on the presentation of logic rules, trees, and programs. It contains sequential derivation based on logic rules \cite{sameer}.  

\begin{figure*}[h]
\includegraphics[scale=0.80]{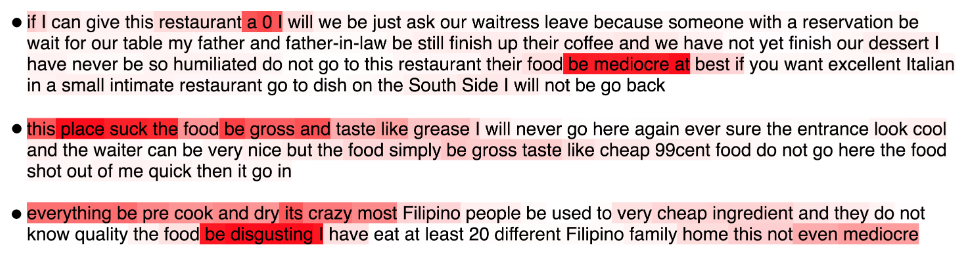}
\caption{ Saliency highlighting \cite{lin}}
\label{visualization1}
\end{figure*}

\begin{figure*}[h]
\includegraphics[scale=1.00]{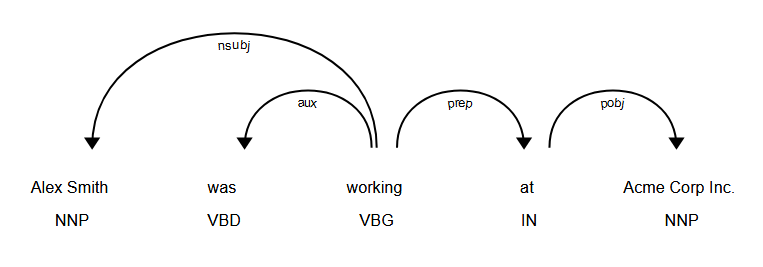}
\caption{ Visualizaton of POS tags\cite{spacy}}
\label{visualization2}
\end{figure*}

\textbf{Natural language explanation:} In this explainable technique, the explanation is provided in more comprehensive natural language\cite{reiter}. It is an application of the generative neural network model where natural language sentences are generated by the NN (Neural network) model \cite{Qiuchi}. For this purpose, sophisticated and dedicated models of particular domains eg. pharma, medical, crime, etc\cite{gen}.  are trained and deployed in production. This model is usually known as the generative model. Fig. \ref{visualization1} and Fig. \ref{visualization2} are describing saliency based highlighting and POS based tags for visualization.

%As shown in Fig. \ref{visualization}, there are many approaches to explain prediction and results in context of natural language processing.

\subsection{XAI and Hate speech detection}
\label{sec:hate speech}
\begin{figure}[h]
\includegraphics[scale=0.45]{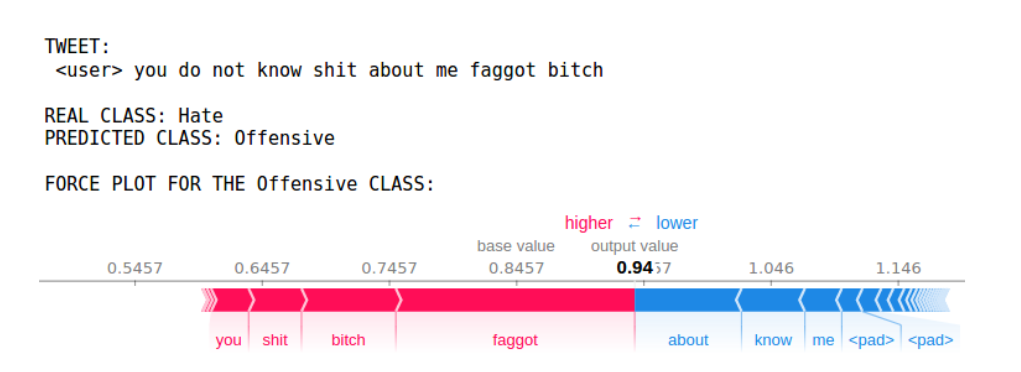}
\caption{Directed  hate  misclassified  as  offensive  language  with  very  high  confidence(94\%).  The two words “bitch” and “faggot” are the two main positive contributors to the score.  Although the two words are indeed offensive,  they misdirect the classifier which misses the clear hate emerging from the tweet  \cite{mosca}.}
\label{hate}
\end{figure}

In this section, we are explaining hate speech detection using XAI. This section demonstrates explainability techniques for  hate speech detection using XAI \cite{duggam} \cite{mathew}. There are text classification based techniques which are useful for providing more insights about trained model for hate speech detection and used data set for training \cite{arras} \cite{chatz} \cite{davidson}.   This projected insights are useful to make trained model more accurate for hate speech detection. Fig. \ref{hate} shows saliency map for classification of hate vs offensive speech. Fig. \ref{hate} shows how certain key words draws classification towards specific class of hate speech with higher weight.

Shapley values are useful for feature importance based analysis of hate speech. Feature importance map like saliency maps are useful for visualization. Gradient explainer based approach is also useful but generate feature independence  based justification.

\section{XAI and VIDEO}
\label{sec:video_sec}
Local optimum justifications are an effective technique for the image domain. A model agnostic technique like LIME shows a good success rate using local explanations. . For video analysis and explainability, frame-wise decomposition of video is applied.

LRP is another popular technique. LRP assumes that it can access the internal architecture of a given complex neural network. LRP access model's internal weight, bias, and activation function for backward propagation.
LRP is structured as a tool for pixel-wise decomposition for relevance to a decision.
LRP satisfies certain conditions for developing justifications as below. 

\begin{itemize}
    \item The relevance difference of each layer must converge must sum at the final layer of the model.
    \item The relevance difference at any neuron of a given layer apart from the final layer is the sum of incoming relevance differences to that layer.
\end{itemize}

Other techniques like LIME are useful to provide justifications to develop explanations for models without accessing internal structure and weights.
LIME approximates decisions by many sampled inputs. LIME is effectively applicable to data like text, image, and tabular data.

The most popular of these techniques, known as Local Interpretable Model Explanations or LIME, seeks to approximate the decision function by many closely sampled input points, which all center around the input point to be explained. It can then attribute positive or negative influence on the decision function to the differences in the sampled in-puts, and overlay this on the original input.

\section{XAI and AUDIO}
\label{sec:audio_sec}
In this area, much research is yet to be done. For linguistic applications, audio waves are converted into text form. After converting into text, NLP-based XAI techniques are applicable. Such techniques comprise lime, perturbation, SHAP and taxonomy-based inductions \cite{shi} \cite{chen}.

Now a days, auto speech recognition powered voice assistant like alexa and siri are being used more frequently by users \cite{preez}. Audio waveform based key word classification for virtual agents is more convincing along with visual presence of agent rather than only voice  or text based output \cite{katharina}.

It is observed that the visual presence of virtual agents in graphical 2d or 3d forms develops trust in the XAI systems. To evaluate this observation, a user study is conducted in which a virtual agent demonstrates XAI visualization of a neural network-based speech recognition model. This model classifies audio keywords with respect to their spectrograms. In this study, users are classified into three groups. First, interact with the text. Second, interact with voice, and Third, interact with virtual agents \cite{katharina}. The results show that the visual appearance of an agent gains more trust rather than only text or voice-based interactions.  

\begin{figure}[h]
\includegraphics[scale=0.50]{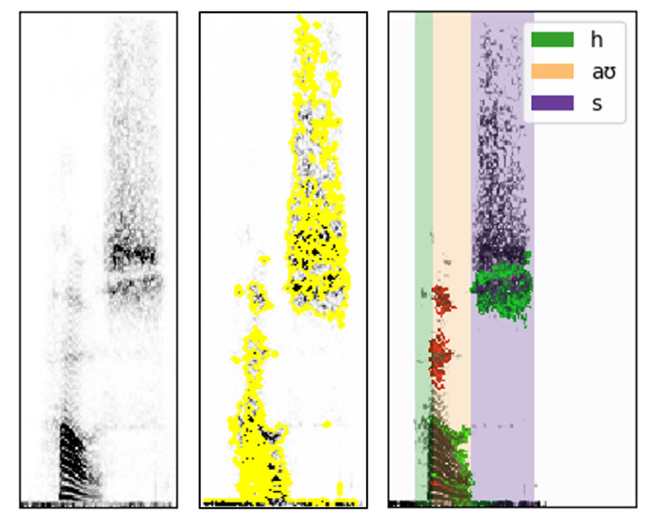}
\caption{A spectrogram of an audio sample (left), its segmentation into superpixels (center) and the output for the user containing LIME visualisations and additional phoneme information (right)\cite{katharina}}
\label{audio}
\end{figure}

LIME framework is applied to generate XAI visualization to understand voice classification. Model agnostic characteristics of LIME make it applicable to any sort of input data. Fig \ref{audio} shows XAI visualization of the keyword "House".

\section{XAI and Multimodal data}
\label{sec:predictive maintenance}
Some time input data may differ from conventional input data like audio, video, text, and image. If data is in CSV format, it requires different pre-processing of data and normalization. For example, in industry sensor data are calibrated in CSV datasheets.

In industrial automation, maintenance is one of the crucial aspects for the continuity of industry. Due to various physical parameters like temperature, vibration, pressure, RPM, etc., there are significant impacts on various parts of the assembly line or mechanical system which leads to failure \cite{bahrudin}.
XAI along with failure diagnosis makes the ML model more transparent and interpretable towards the provided diagnosis of the failed component \cite{shukla}.

In the aviation sector, aircraft maintenance is handled by scheduled or event-based triggered maintenance. Such sort of maintenance is unreliable because that causes serious disaster when aircraft is in the air. Such disasters can be prevented if predictive maintenance is applied. In any giant mechanical system, there is a gradual degradation in the reading of various tools or sensors, these tools are making a cumulative effect for final break down \cite{matka}. Such calibration of readings can be used as a feature set and model training is possible for failure diagnosis and remaining useful life (RUL)
prediction of aircraft or any critical electro-mechanical system.

The prediction of failure with explanation makes the justification for derived diagnosis. Hence it improves reliability and saves cost\cite{sophie}.

\begin{figure*}[h]
\includegraphics[scale=0.5]{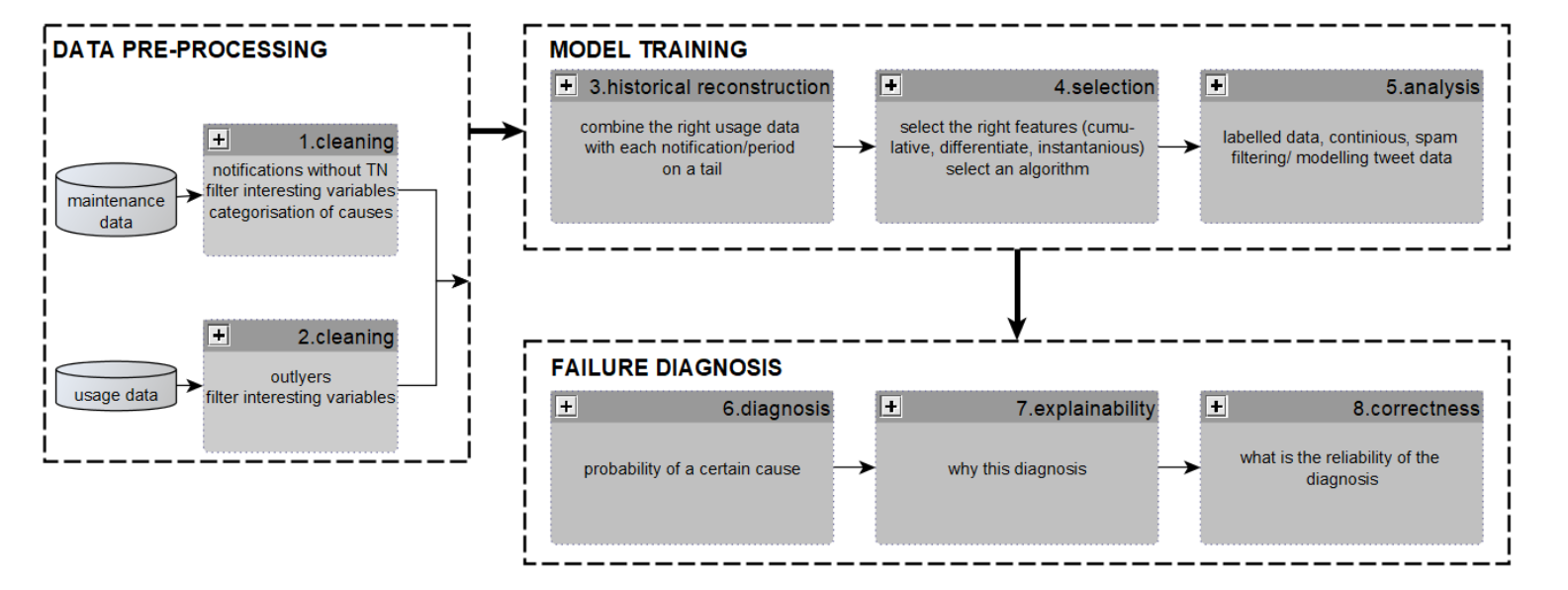}
\caption{Functional diagram of an automated failure diagnosis model \cite{sophie}}
\label{failure}
\end{figure*}

Maintenance with interpretability for failure diagnosis can add useful insights about the disposal of certain parts which might not be available with crew members of maintenance teams. The pipeline of predictive maintenance with XAI insights is mentioned in Fig. \ref{failure}. of sequential steps like data collection, data cleaning, feature selection, diagnosis, and explanation with validation. One advantage of predictive maintenance it helps to mandatory understanding of different physical components and their physical properties.

\section{Conclusion}
It is emphasized here that XAI is an important and mandatory aspect of AI/ML based application to use in real time. Our study has started discussion from conventional AI and limitations. The need for XAI is well explained in the case of studies along with key issues of explainable AI. 

Objectives and scopes of XAI are discussed in length and breadth. We discussed major objectives like transparency, fairness, bias, and confidence. Scope of XAI is discussed in detail for its application in the major domain like NLP, medical, defense and Engineering.

Different methodologies (post hoc and transparent)  for explainability are discussed to get preliminary hands on to dive into this field. Conceptual and detailed explanations with the example for all methodologies are also discussed. After providing a conceptual understanding of XAI approaches, we have provided XAI as a tool to be applied to specific kinds of data like image, text, video, audio, and multimodal data.

This survey elaborates a conceptual understanding of XAI along with the importance of explainability that motivates researchers for diversified aspects of XAI. This purpose motivates researchers for interpretable AI/ML methods. These detailed highlights make a baseline for the understanding of the current literature of XAI, which can be approached in two ways. 1) Transparent ML models which are interpretable to an extent by themselves only. 2) Post hoc methods for explainability which makes the model more interpretable. We presented XAI as a tool for responsible AI, a paradigm that can enable series of algorithms that will work in synergy to achieve the goal of responsible AI. Responsible AI stands for trust, confidence, fairness, and transparency.

\EOD


\begin{thebibliography}{10}

 \bibitem[1]{Adadi} A. Adadi and M. Berrada, "Peeking Inside the Black-Box: A Survey on Explainable Artificial Intelligence (XAI)," in IEEE Access, vol. 6, pp. 52138-52160, 2018, doi: 10.1109/ACCESS.2018.2870052.
 
 \bibitem[2]{marco} Marco Tulio Ribeiro,Sameer Singh, Carlos Guestrin,“Why Should I Trust You?”Explaining the Predictions of Any Classifier," https://arxiv.org/pdf/1602.04938.pdf

 
 \bibitem[3]{SHAP} Scott M Lundberg and Su-In Lee. A unified approach to interpreting model predictions. In Advances in Neural Information Processing Systems, pages 4765–4774, 2017.
 
 \bibitem[4]{LRP} Sebastian Bach, Alexander Binder, Grégoire Montavon, Frederick Klauschen, Klaus-Robert Müller, and Wojciech Samek. On pixel-wise explanations for non-linear classifier decisions by layer-wise relevance propagation. PloS one, 10(7):e0130140, 2015.
 
\bibitem[5]{sparse} J. Wang, J. Yang, K. Yu, F. Lv, T. Huang and Y. Gong, "Locality-constrained Linear Coding for image classification," 2010 IEEE Computer Society Conference on Computer Vision and Pattern Recognition, San Francisco, CA, 2010, pp. 3360-3367, doi: 10.1109/CVPR.2010.5540018.

\bibitem[6]{shap2}Lundberg, Scott M., Gabriel G. Erion, and Su-In Lee. "Consistent individualized feature attribution for tree ensembles." arXiv preprint arXiv:1802.03888 (2018).

\bibitem[7]{dhurandhar}Amit Dhurandhar, Pin-Yu Chen, Ronny Luss, Chun-Chen Tu, Paishun Ting, Karthikeyan Shanmugam, and Payel Das. Explanations based on the missing: Towards contrastive explanations with pertinent negatives. In Advances in Neural Information Processing Systems, pages 592–603, 2018

\bibitem[8]{tom}Vermeire, Tom and D. Martens. “Explainable Image Classification with Evidence Counterfactual.” ArXiv abs/2004.07511 (2020): n. pag.

\bibitem[9]{lloyd}Lloyd S Shapley. “A value for n-person games”. In:Contributions to the Theory of Games2.28 (1953), pp. 307–317

\bibitem[10]{turing} Vaishak Belle Ioannis Papantonis.,Principles and Practice of Explainable Machine Learning*, Sep 2020

\bibitem[11]{rudin} Cynthia Rudin, Stop explaining black box machine learning models for high stakes decisions and use
interpretable models instead,


\bibitem[12]{igami} Igami. (2017). ‘‘Artificial intelligence as structural estimation: Eco-nomic interpretations of deep blue, bonanza, and AlphaGo.’’

\bibitem[13]{neer} A. Neerincx, J. van der Waa, F. Kaptein, and J. van Diggelen, ‘‘Using perceptual and cognitive explanations for enhanced human-agent team performance,’’ in Proc. Int. Conf. Eng. Psychol. Cogn. Ergonom. (EPCE),2018, pp. 204–214.

\bibitem[14]{garcia} J. C. Garcia, D. A. Robb, X. Liu, A. Laskov, P. Patron, and H. Hastie,‘‘Explain yourself: A natural language interface for scrutable autonomous robots,’’ inProc. Explainable Robot. Syst. Workshop HRI, 2018

\bibitem[15]{Mantong} Mantong  Zhou,   Minlie  Huang,   and  Xiaoyan  Zhu.2018.  An interpretable reasoning network for multi-relation  question  answering.In Proceedings  of the 27th International Conference on Computational Linguistics, 

\bibitem[16]{Qizhe} Qizhe Xie, Xuezhe Ma, Zihang Dai, and Eduard Hovy.2017.   An  interpretable  knowledge  transfer  model for  knowledge  base  completion.In Proceedings of  the  55 th  Annual  Meeting  of  the  Association  for Computational Linguistics (Volume 1: Long Papers),pages 950–962, Vancouver, Canada. Association for Computational Linguistics

\bibitem[17]{Nikos}Nikos   Voskarides,   Edgar   Meij,   Manos   Tsagkias,Maarten  de  Rijke,  and  Wouter  Weerkamp.  2015.Learning  to  explain  entity  relationships  in  knowledge  graphs.    In Proceedings  of  the  53rd  Annual Meeting of the Association for Computational Linguistics and the 7th International Joint Conferenceon Natural Language Processing

\bibitem[18]{Attention} Ashish  Vaswani,  Noam  Shazeer,  Niki  Parmar,  JakobUszkoreit,  Llion  Jones,  Aidan  N.  Gomez,  ŁukaszKaiser, and Illia Polosukhin. 2017.  Attention is all you need. In NeuralIPS

\bibitem[19]{Martin} Martin  Tutek  and  JanˇSnajder.  2018.   Iterative  recursive attention model for interpretable sequence classification. In Proceedings of the 2018 EMNLP Work-shop BlackboxNLP: Analyzing and Interpreting Neural Networks for NLP, Brussels, Belgium. Association for Computational Linguistics.

\bibitem[20]{james} James Thorne,Andreas Vlachos,Christos Christodoulopoulos, and Arpit Mittal. 2019.  Gener-ating token-level explanations for natural language inference. In Proceedings of the 2019 Conference of the North American Chapter of the Association for Computational Linguistics: Human Language Technologies, Volume 1 (Long and Short Papers),  Minneapolis,  Minnesota. Association for Computational Linguistics.

\bibitem[21]{Robert}Robert Schwarzenberg, David Harbecke, Vivien Mack-etanz, Eleftherios Avramidis, and Sebastian  Moller.2019.    Train,  sort,  explain:   Learning  to  diagnose translation models. In Proceedings of the 2019 Conference of the North American Chapter of the Association  for  Computational  Linguistics  (Demonstrations)

\bibitem[22] {Sofia} Sofia Serrano and Noah A. Smith. 2019.   Is attention interpretable ? In Proceedings of the 57th Annual Meeting of the Association for Computational Linguistics, Florence, Italy. Association for Computational Linguistics.

\bibitem[23]{Karen} Karen Simonyan, Andrea Vedaldi, and Andrew Zisserman. 2013. Deep inside convolutional networks: Visualising  image  classification  models  and  saliency maps. arXiv preprint arXiv:1312.6034

\bibitem[24]{Nina} Nina  Poerner,  Hinrich  Schutze,  and  Benjamin  Roth.2018.  Evaluating neural network explanation methods  using  hybrid  documents  and  morpho syntactic agreement. In Proceedings of the 56th Annual Meeting  of  the  Association  for  Computational  Linguistics (Volume 1:  Long Papers), Melbourne,  Australia.  Association  for  Computational Linguistics.

\bibitem[25]{Nicolas} Nicolas  Prollochs,  Stefan  Feuerriegel,  and  Dirk  Neumann. 2019.   Learning interpretable negation rules via weak supervision at document level: A reinforcement learning approach. In Proceedings of the 2019Conference  of  the  North  American  Chapter  of  the Association for Computational Linguistics:  Human Language Technologies, Volume 1 (Long and Short Papers),  Minneapolis,  Minnesota.Association for Computational Linguistics.


\bibitem[26]{Nazneen} Nazneen  Fatema  Rajani,   Bryan  McCann,   Caiming Xiong,  and  Richard  Socher.  2019b.   Explain  your-self!  leveraging language models for common sense reasoning.arXiv preprint arXiv:1906.02361.

\bibitem[27]{Reid} Reid  Pryzant,  Sugato  Basu,  and  Kazoo  Sone.  2018a. Interpretable neural architectures for attributing anad’s performance to its writing style. In Proceedings of  the  2018  EMNLP  Workshop  Blackbox NLP:  Analyzing and Interpreting Neural Networks for NLP, Brussels, Belgium. Association for Computational Linguistics.

\bibitem[28]{Piyawat} Piyawat Lertvittayakumjorn and Francesca Toni. 2019.Human-grounded  evaluations  of  explanation  methods  for  text  classification.    In Proceedings  of  the 2019  Conference  on  Empirical  Methods  in  Natural Language Processing and the 9th International Joint Conference on Natural Language Processing (EMNLP-IJCNLP). Hong Kong,China. Association for Computational Linguistics.

\bibitem[29]{Jiwei} Jiwei Li, Xinlei Chen, Eduard Hovy, and Dan Jurafsky. 2015.  Visualizing and understanding neural models in nlp.arXiv preprint arXiv:1506.01066.

\bibitem[30]{Qiuchi} Qiuchi Li, Benyou Wang, and Massimo Melucci. 2019.CNM: An interpretable complex valued network for matching.   In Proceedings of the 2019 Conference of  the  North  American  Chapter  of  the  Association for  Computational  Linguistics:   Human  Language Technologies,  Volume  1  (Long  and  Short  Papers),pages 4139–4148, Minneapolis, Minnesota. Association for Computational Linguistics.

\bibitem[31]{Vijay} Vijay N. Garla, Cynthia Brandt,
Ontology-guided feature engineering for clinical text classification,
Journal of Biomedical Informatics, Volume 45, Issue 5,2012, Pages 992-998, ISSN 1532-0464, https://doi.org/10.1016/j.jbi.2012.04.010.

\bibitem[32]{Bahdanau} Minh-Thang Luong, Hieu Pham, Christopher D. Manning, Effective Approaches to Attention-based Neural Machine Translation

\bibitem[33]{Ashish} Ashish Sureka, Pankaj Jalote ,Detecting Duplicate Bug Report Using Character N-Gram-Based Features

\bibitem[34]{Prihatini} P M Prihatini, I K Suryawan, IN Mandia. Feature extraction for document text using Latent Dirichlet Allocation

\bibitem[35]{abuj} Abdalghani Abujabal, Mohamed Yahya, Mirek Riedewald, and Gerhard Weikum. 2017. Automated Template Generation for Question Answering over Knowledge Graphs. In Proceedings of the 26th International Conference on World Wide Web (WWW '17). International World Wide Web Conferences Steering Committee, Republic and Canton of Geneva, CHE, 1191–1200. DOI:https://doi.org/10.1145/3038912.3052583

\bibitem[36]{mahnaz} Mahnaz Koupaee, William Yang Wang, IN Mandia. Analyzing and Interpreting Convolutional Neural Networks in NLP.



\bibitem[37]{lin} Lin, Zhouhan, et al. "A structured self-attentive sentence embedding." arXiv preprint arXiv:1703.03130 (2017).

\bibitem[38]{reiter} REITER, E., \& DALE, R. (1997). Building applied natural language generation systems. Natural Language Engineering, 3(1), 57-87. doi:10.1017/S1351324997001502

\bibitem[39]{gen} Dani Yogatama, Chris Dyer, Wang Ling, and Phil Blunsom, deep mind, Generative and Discriminative Text Classificationwith Recurrent Neural Networks

\bibitem[40]{sameer} Dani Yogatama, Chris Dyer, Wang Ling, and gren, H. \& Nieves, J.C. A dialogue-based approach for dealing with uncertain and conflicting information in medical diagnosis. Auton Agent Multi-Agent Syst 32, 861–885 (2018). https://doi.org/10.1007/s10458-018-9396-x

\bibitem[41]{taylor} Montavon, G.; Lapuschkin, S.; Binder, A.; Samek, W.; Müller, Phil Blunsom, deep mind, Generative and Discriminative Text Classificationwith Recurrent Neural Networks

\bibitem[42]{singh} Amitojdeep Singh, Sourya Sengupta,Vasudevan Lakshminarayanan, Explainable Deep Learning Models in Medical Image Analysis,https://doi.org/10.3390/jimaging6060052


\bibitem[43]{yan} Yan, C., LindK.R. Explaining nonlinear classification
decisions with deep taylor decomposition. Pattern Recognit. 2017, 65, 211–222.

\bibitem[44]{ancona}Ancona, M.; Ceolini, E.; Öztireli, C.; Gross, M. Towards better understanding of gradient-based attribution
methods for deep neural networks. arXiv 2017, arXiv:1711.06104.


\bibitem[45]{avanti} Avanti Shrikumar, Peyton Greenside, Anshul Kundaje,Learning Important Features Through Propagating Activation Differences. https://arxiv.org/abs/1704.02685

\bibitem[46]{erico} Erico Tjoa, and Cuntai Guan,Fellow, IEEE, A Survey on Explainable Artificial Intelligence(XAI): towards Medical XAI. https://arxiv.org/pdf/1907.07374.pdf

\bibitem[47]{liam} Liam  Hiley,  Alun  Preece,  Yulia  Hicks,  Supriyo  Chakraborty,  Prudhvi Gurram, and Richard Tomsett. Explaining motion relevance for activity recognition in video deep learning models, 2020

\bibitem[48]{doshi} Doshi-Velez  and  Been  Kim.    Towards  a  rigorous  science  of interpretable machine learning, 2017.  cite arxiv:1702.0860

\bibitem[49]{sebastian} Sebastian Lapuschkin, Stephan Waldchen, Alexander Binder, Gregoir eMontavon,  Wojciech  Samek,  and  Klaus-Robert  Muller.    Unmasking clever hans predictors and assessing what machines really learn.Nature Communications, 10(1):1096, 2019.

\bibitem[50]{gilpin} L. H. Gilpin, D. Bau, B. Z. Yuan, A. Bajwa, M. Specter, and L. Kagal. Explaining explanations:  An  overview  of  interpretability  of  machine learning.  In 2018 IEEE 5th International Conference on Data Science and Advanced Analytics (DSAA), pages 80–89, 2018


\bibitem[51]{fernandez} A. Fernandez, F. Herrera, O. Cordon, M. Jose del Jesus, and F. Marcel-loni.  Evolutionary fuzzy systems for explainable artificial intelligence:Why, when, what for, and where to?IEEE Computational Intelligence Magazine, 14(1):69–81, Feb 2019

\bibitem[52]{goire} Grgoire Montavon, Wojciech Samek, and Klaus-Robert Mller. Methodsfor interpreting and understanding deep neural networks.Digital Signal Processing, 73:1 – 15, 2018.

\bibitem[53]{mosca}Edoardo Mosca,Explainability of Hate Speech Detection Models, Technische Universit at Munchen, Department of Mathematics

\bibitem[54]{arras} Arras, L., Montavon, G., Muller,  \& Samek, W. (2017). Explaining recurrent neural  network  predictions  in  sentiment  analysis.  In Proceedings  of  the  8th  workshop on  computational  approaches  to  subjectivity,  sentiment  and  social  media  analysis

\bibitem[55]{pelekis} Baziotis, C., Pelekis, N., \& Doulkeridis, C. (2017). Datastories at semeval-2017 task 4:Deep lstm with attention for message-level and topic-based sentiment analysis. In Proceedings  of  the  11th  international  workshop  on  semantic  evaluation.  (semeval-2017)(pp. 747–754)

\bibitem[56]{chen} Chen,  Y.,  Zhou,  Y.,  Zhu,  S.,  \&  Xu,  H.  (2012).  Detecting  offensive  language  in  social media to protect adolescent online safety. In2012 international conference on privacy, security, risk and trust and 2012 international conference on social computing


\bibitem[57]{chatz} Chatzakou, D., Kourtellis, N., Blackburn, J., De Cristofaro, E., Stringhini, G., \& Vakali,A. (2017). Mean birds: Detecting aggression and bullying on twitter. In Proceedings of the 2017 acm on web science conference(pp. 13–22).

\bibitem[58]{davidson} Davidson, T., Warmsley, D., Macy, M., \& Weber, I. (2017). Automated hate speech detection and the problem of offensive language. In Eleventh international aaai conference on web and social media

\bibitem[59]{dixon} Dixon, L., Li, J., Sorensen, J., Thain, N., \& Vasserman, L. (2018). Measuring and mitigating unintended bias in text classification. In Proceedings  of  the  2018  aaai/acm conference on ai, ethics, and society(pp. 67–73).

\bibitem[60]{duggam} Online harassment 2017. Pew Research Center

\bibitem[61]{founta}Founta, A. M., Chatzakou, D., Kourtellis, N., Blackburn, J., Vakali, A., \& Leontiadis, I.(2019). A unified deep learning architecture for abuse detection. In Proceedings  of the 10 th acm conference on web science(pp. 105–114).

\bibitem[62]{glorot}Glorot, X., Bordes, A., \& Bengio, Y. (2011). Domain adaptation for large-scale sentiment classification:  A  deep  learning  approach.  In Proceedings  of  the  28th  international conference on international conference on machine learning(513–520).

\bibitem[63]{mathew} Mathew, B., Dutt, R., Goyal, P., \& Mukherjee, A. (2019). Spread of hate speech in onlinesocial media. InProceedings  of  the  10th  acm  conference  on  web  science(pp. 173–182).

\bibitem[64]{eitel} Eitel, F.; Ritter, K.; Alzheimer’s Disease Neuro imaging Initiative (ADNI). Testing the Robustness of Attribution Methods for Convolutional Neural Networks in MRI-Based Alzheimer’s Disease Classification. In Interpretability of Machine Intelligence in Medical Image Computing and Multimodal Learning for Clinical
Decision Support, ML-CDS 2019, IMIMIC 2019; Lecture Notes in Computer Science; Suzuki, K., et al., Eds.; Springer: Cham, Switzerland, 2019; Volume 11797.

\bibitem[65]{pereira} Pereira, S.; Meier, R.; Alves, V.; Reyes, M.; Silva, C.A. Automatic brain tumor grading from MRI data using convolutional neural networks and quality assessment. In Understanding and Interpreting Machine Learning in
Medical Image Computing Applications; Springer: Cham, Switzerland, 2018; pp. 106–114.

\bibitem[66]{matteo} Matteo Pennisi, Isaak Kavasidis, Concetto Spampinato, Vincenzo Schininà, Simone Palazzo, Francesco Rundo, Massimo Cristofaro, Paolo Campioni, Elisa Pianura, Federica Di Stefano, Ada Petrone, Fabrizio Albarello, Giuseppe Ippolito, Salvatore Cuzzocrea, Sabrina Conoci, An Explainable AI System for Automated COVID-19 Assessment and Lesion Categorization from CT-scans, https://arxiv.org/abs/2101.11943

\bibitem[67]{sophie} MSophie ten Zeldam, Arjan de Jong, Richard Loenders loot and Tiedo Tinga, Automated Failure Diagnosis in Aviation Maintenance UsingeXplainable Artificial Intelligence (XAI)

\bibitem[68]{mashrur} Mashrur Chowdhury, Adel W. Sadek, Advantages and Limitations of Artificial Intelligence.

\bibitem[69]{rigano} Christopher Rigano, NIJ, USING ARTIFICIAL INTELLIGENCE TO ADDRESS CRIMINAL JUSTICE NEEDS.

\bibitem[70]{robbins}Robbins, Mark D.. (2019). AI Explainability Regulations and Responsibilities. 

\bibitem[71]{zhong}Y. X. Zhong, "A Cognitive Approach to Artificial Intelligence Research," 2006 5th IEEE International Conference on Cognitive Informatics, Beijing, China, 2006, pp. 90-100, doi: 10.1109/COGINF.2006.365682.


\bibitem[72]{das} Arun Das,Graduate Student Member, IEEE,and Paul Rad,Senior Member, IEEE,Opportunities and Challenges in ExplainableArtificial Intelligence (XAI): A Survey",doi : https://arxiv.org/pdf/2006.11371.pdf

\bibitem[73]{lime} "Why Should I Trust You?": Explaining the Predictions of Any Classifier,Marco Tulio Ribeiro, Sameer Singh, Carlos Guestrin, https://arxiv.org/abs/1602.04938

\bibitem[74]{katharina}“Let me explain!”: exploring the potential of virtual agents inexplainable AI interaction design, Katharina Weitz,Dominik Schiller,Ruben Schlagowski,Tobias Huber, Elisabeth Andre 

\bibitem[75]{bahrudin}Hrnjica, Bahrudin \& Softic, Selver. (2020). Explainable AI in Manufacturing: A Predictive Maintenance Case Study. 

\bibitem[76]{shukla} Shukla, Bibhudhendu \& Fan, Ip-Shing \& Jennions, I.K.. (2020). 
Opportunities for Explainable Artificial Intelligence in Aerospace Predictive Maintenance.  

\bibitem[77]{matka} S. Matzka, "Explainable Artificial Intelligence for Predictive Maintenance Applications," 2020, (AI4I), doi: 10.1109/AI4I49448.2020.00023

\bibitem[78]{preez} S. J. du Preez, M. Lall and S. Sinha, "An intelligent web-based voice chat bot," IEEE EUROCON 2009, 2009, pp. 386-391, doi: 10.1109/EURCON.2009.5167660.

\bibitem[79]{shi} Nuobei Shi, Qin Zeng, Raymond Lee,The design and implementation of Language Learning Chatbot with XAI using Ontology and Transfer Learning, (NLPD 2020),doi: https://arxiv.org/abs/2009.13984

\bibitem[80]{Siham} Alejandro Barredo Arrieta et al.,Explainable Artificial Intelligence (XAI): Concepts, Taxonomies, Opportunities and Challenges toward Responsible AI, DOI: 10.1016/j.inffus.2019.12.012.

\bibitem[81]{spacy} visualization to output the fine-grained part-of-speech tags, https://spacy.io/usage/rule-based-matching
\end{thebibliography}
\end{document}